\documentclass[10pt,twocolumn,letterpaper]{article}

\usepackage[pagenumbers]{iccv} 
\usepackage{indentfirst}
\usepackage{makecell}
\usepackage{booktabs}
\usepackage{multirow}
\usepackage{colortbl}
\usepackage{xcolor}
\usepackage{array}
\usepackage{float}

\usepackage{graphicx}	
\usepackage{amsmath}	
\usepackage{amssymb}	
\usepackage{booktabs}
\usepackage{times}
\usepackage{microtype}
\usepackage{epsfig}
\usepackage{caption}
\usepackage{float}
\usepackage{placeins}
\usepackage{color, colortbl}
\usepackage{stfloats}
\usepackage{enumitem}
\usepackage{tabularx}
\usepackage{xstring}
\usepackage{multirow}
\usepackage{xspace}
\usepackage{url}
\usepackage{subcaption}
\usepackage{xcolor}
\usepackage[hang,flushmargin]{footmisc}

\usepackage{graphicx} 
\usepackage{relsize}
\usepackage{amsmath}
\usepackage{caption}
\usepackage{setspace}
\usepackage{float}

\definecolor{iccvblue}{rgb}{0.21,0.49,0.74}
\usepackage[pagebackref,breaklinks,colorlinks,allcolors=iccvblue]{hyperref}

\def\paperID{6026} 
\def\confName{ICCV}
\def\confYear{2025}

\title{LightBSR: Towards Lightweight Blind Super-Resolution via Discriminative Implicit Degradation Representation Learning}

\author{
Jiang Yuan$^{1,2,4}$\footnotemark[1], Ji Ma$^{1,2}$\footnotemark[1], Bo Wang$^1$\footnotemark[2], Guanzhou Ke$^{3}$, Weiming Hu$^1$\\
{\hspace{-0.4cm}\normalsize $^1$State Key Laboratory of Multimodal Artificial Intelligence Systems, Institute of Automation, Chinese Academy of Sciences,}  \\
{\hspace{-0.4cm}\normalsize $^2$North China Electric Power University, $^3$Beijing Jiaotong University, $^4$YunQue AGI, Hangzhou, China}  \\
{\tt\small \
\{yuanj, maji\}@ncepu.edu.cn, guanzhouk@gmail.com, wangbo@ia.ac.cn, wmhu@nlpr.ia.ac.cn}
}

\begin{document}
\maketitle

\renewcommand{\thefootnote}{\fnsymbol{footnote}}
\footnotetext[1]{Authors contributed equally. Work done during the internship at CASIA.}
\footnotetext[2]{Corresponding author.}

\begin{abstract}
Implicit degradation estimation-based blind super-resolution (IDE-BSR) hinges on extracting the implicit degradation representation (IDR) of the LR image and adapting it to LR image features to guide HR detail restoration.
Although IDE-BSR has shown potential in dealing with noise interference and complex degradations, 
existing methods ignore the importance of IDR discriminability for BSR and instead over-complicate the adaptation process to improve effect, resulting in a significant increase in the model's parameters and computations.
In this paper, we focus on the discriminability optimization of IDR and propose a new powerful and lightweight BSR model termed LightBSR.
Specifically, we employ a knowledge distillation-based learning framework.
We first introduce a well-designed degradation-prior-constrained contrastive learning technique during teacher stage to make the model more focused on distinguishing different degradation types.
Then we utilize a feature alignment technique to transfer the degradation-related knowledge acquired by the teacher to the student for practical inferencing.
Extensive experiments demonstrate the effectiveness of IDR discriminability-driven BSR model design.
The proposed LightBSR can achieve outstanding performance with minimal complexity across a range of blind SR tasks.
Our code is accessible at: \url{https://github.com/MJ-NCEPU/LightBSR}.
\end{abstract}
\vspace{-1.0em}    
\section{Introduction}
\label{sec:intro}
\vspace{-4pt}
The goal of blind super-resolution (BSR) \cite{cycle-consistent,ikc_14,DCLS,DKP,DARSR,dasr_5,daa,PL-IDENet,IDEnet,mrda_9,kdsr,CDFormer} is to restore the detail-rich high-resolution (HR) image from its low-resolution (LR) counterpart with unknown and complex degradations.
The key to BSR methods lies in accurately estimating the degradation information of the LR image and embedding it into the SR module effectively to guide image upsampling.
As a more practical LR image restoration technique, BSR has attracted a lot of attention and developed two research directions: explicit degradation estimation-based (EDE-) BSR and implicit degradation estimation-based (IDE-) BSR.\par
\begin{figure}[t]
      \centering
      \includegraphics[width=0.99\linewidth]{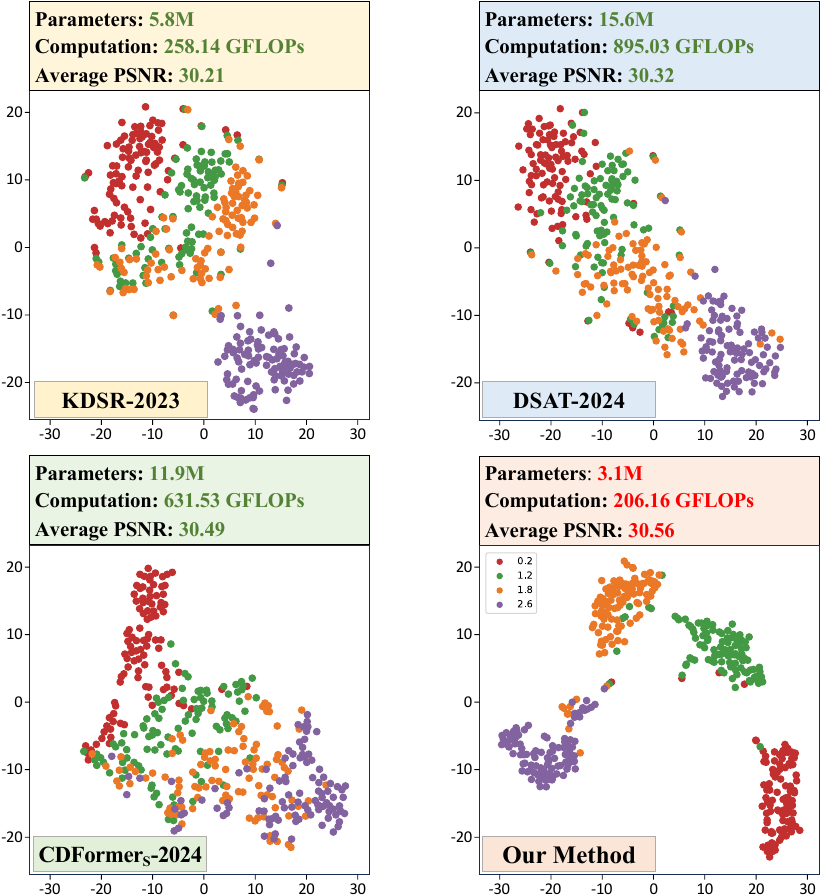}\vspace{-5pt}
      \caption{
      Comparison of different methods in terms of
      model complexity, 
      SR performance and IDR distribution across four blur kernels on the DIV2K \cite{div2k_33} val-set.           
      }
      \vspace{-18pt}
      \label{fig:illusion}
\end{figure}

The EDE-BSR paradigm \cite{kernelgan,koalanet_23,bsrgan,ikc_14,dan_13,DCLS} requires an explicit mapping between degradation parameters (e.g., blur kernel, noise level) and the estimator. Only when the degradation type of the LR image falls within the predefined degradation scope can the EDE-BSR method achieve satisfactory image restoration results.
However, real-world degradation scenarios are nearly infinite, making it impractical to annotate all possible types of degradation in advance.
Additionally, precise kernel estimation often necessitates more parameters and iterations \cite{ikc_14,dan_13}, significantly increasing the model and time complexity of such methods.\par
The IDE-BSR paradigm \cite{dasr_5,idmbsr_7,mrda_9,kdsr,ASSR,DSAT,CDFormer} transforms the problem of estimating degradation information into a degradation-related representation learning problem. 
The core objective is to construct a latent representation space that can generalize to various degradation domains, including unseen types during training, 
and effectively adapt the extracted implicit degradation representation (IDR) to LR image features to guide HR image restoration.
Benefiting from its more comprehensive understanding of the intrinsic characteristics of intricate degradation patterns and simpler training process, IDE-BSR paradigm has shown significant potential in real-world scenarios with noise interference and complex degradation combinations.\par
The previous works have explored the application of various representation learning techniques in IDE-BSR task, such as contrastive learning \cite{mocov1,dasr_5,DSAT}, meta learning \cite{meta-learning,mrda_9}, distillation learning \cite{kd,kdsr}, and diffusion model \cite{diffusion,CDFormer}. 
Despite some progress, these methods did not particularly emphasize the discriminative power of the learned IDR space.
As shown in the t-SNE plots of Fig.\ref{fig:illusion}, even the latest methods \cite{DSAT,CDFormer} struggle to distinguish different degradation types effectively. 
This means that these models do not accurately understand different degradation patterns, and the extracted IDR contains degradation-independent noise.
As a result, they have to increase the complexity of the adaptation process to solve the issue of noise interference for improving SR effect.
For example, although CDFormer\textsubscript{S} \cite{CDFormer} and DSAT \cite{DSAT} achieved high average PSNR values on the DIV2K val-set \cite{div2k_33}, the cost was an increase of more than twice the number of parameters and computational costs compared to KDSR \cite{kdsr}. 
However, the increasing complexity of models undoubtedly hinders the practicability of BSR technology.
This phenomenon raises an important question: \textit{is it possible to enhance SR effect by strengthening IDR discriminability, thereby circumventing complex adaptation processes and achieving a lightweight BSR model?}
\par
\begin{figure*}[t] {\centering
\centerline{\includegraphics[width=0.99\linewidth]{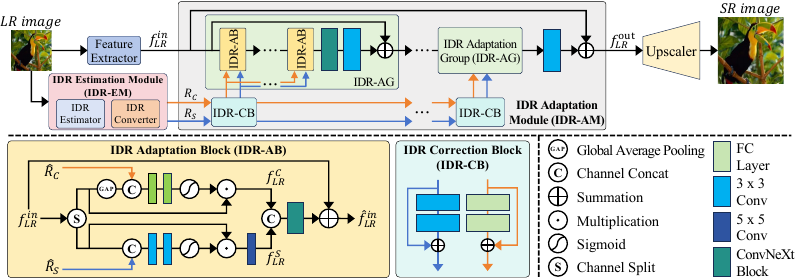}}\vspace{-2pt}
      \caption{
      The architecture of our designed implicit degradation estimation-based lightweight blind SR network.
      }
      \vspace{-14pt}
      \label{fig:overall}
      }
\end{figure*}

%
Based on the above observations and reflections, we attempt to focus on enhancing the discriminability of IDRs rather than stacking adaptation networks in this paper, aiming to form a new powerful and parameter-efficient BSR model design.
The proposed method is termed \textbf{LightBSR}.
The overall architecture is shown in Fig.\ref{fig:overall}, where the core modules are an IDR estimation module (IDR-EM) and an IDR adaptation module (IDR-AM).
The former consists of an IDR-estimator and an IDR-converter, used to extract channel- and spatial-wise IDRs from the input LR image.
The latter consists of a series of hierarchical adaptation components, used to effectively adapt IDRs to LR image features.
Regarding model learning, as shown in Fig.\ref{fig:contrast}, we follow the distillation learning framework that has a natural advantage in lightweight, allowing us to perform very complex implicit degradation modeling in the teacher stage without affecting the model complexity and inference efficiency in the student stage.
During teacher training stage,
we introduce a well-designed degradation-prior-constrained contrastive learning technique, which uses specific degradation parameters as degradation reference priors (DRP) to reinforce the teacher's learning of discriminative degradation-related information in LR images.
In the student training stage,
we utilize a feature alignment-based distillation technique to transfer the degradation-related knowledge acquired by the teacher to a simplified student model that only takes the LR image as input to meet the task requirement.
\par
Extensive experiments on both synthetic and real-world images demonstrate the effectiveness of our novel BSR model design.
Thanks to the high discriminability of IDRs, even with a low complexity degradation adaptation module, it is easy to build a high-quality mapping from LR to HR with the generalization to any degradation scenarios.
Compared to existing advanced EDE- and IDE-BSR methods under the same settings, our method achieves outstanding performance across various blind SR tasks with minimal parameters and computational costs.

\vspace{-6pt}
\section{Related Work}
\label{sec:related}
\vspace{-5pt}

The BSR aims to extract degradation information from LR images using a learnable estimator instead of manually set parameters \cite{srmd_20,udvd}, to guide reconstruction. \textbf{EDE-BSR.} Early methods explicitly estimate degradation. IKC \cite{ikc_14} iteratively refines the estimator with generated SR results. DAN \cite{dan_13} jointly predicts blur kernels and SR images in a dual-branch design. DCLS \cite{DCLS} introduces constrained least-square filtering to produce deblurred features. These approaches often involve costly iterative estimation. \textbf{IDE-BSR.} Recent works model degradation implicitly by learning latent representations integrated with LR features. DASR \cite{dasr_5} first applied contrastive learning \cite{mocov1} to distinguish degradations. IDMBSR \cite{idmbsr_7} uses kernel width and noise level as weak supervision. MRDA \cite{mrda_9} adopts meta-learning \cite{meta-learning} in a multi-stage setup. KDSR \cite{kdsr} introduces knowledge distillation \cite{kd,fitnets,ke2024rethinking}, transferring knowledge from HR-supervised teachers to students for degradation estimation. CDFormer \cite{CDFormer} and DSAT \cite{DSAT} stack large Transformer blocks \cite{swinir_15,DAT} for performance gains, but their complexity limits practical use. \textbf{Our Goal.} Instead of enlarging models, we enhance the discriminability of the IDR space by combining contrastive learning and distillation, enabling lightweight yet effective BSR. Further analysis is in the supplementary material.

\section{Method}
\label{sec:method}

\subsection{Architecture Overview}\label{overviwer}
\vspace{-3pt}
The architecture of the proposed LightBSR is shown in Fig.\ref{fig:overall}, which mainly consists of an LR Feature Extractor (a 3$\times$3 Conv layer), an IDR Estimation Module (IDR-EM), an IDR Adaptation Module (IDR-AM) and an Upscaler \cite{pixshuffle_49}.
The LR image is first fed into both the low-level LR image feature extractor and IDR-EM simultaneously.
Secondly, outputs are fed into IDR-AM for a series of modulation and fusion.
Thirdly, the adapted result is fed into the Upscaler to generate the SR image.
More design details are below.
\par
\vspace{-3pt}
\subsubsection{IDR Estimation Module (IDR-EM)}
IDR-EM consists of an IDR-estimator and an IDR-converter.
The former is used to extract the original IDR, and its structure is a six-layer Conv network like KDSR \cite{kdsr}, but halves channels of each layer to reduce parameters.
The latter includes a spatial transformation branch (consisting of a pixel-shuffle layer and a Conv layer) and a channel transformation branch (consisting of a GAP operator and a FC layer) for further generating spatial-wise IDR and channel-wise IDR based on original IDR, respectively.
\vspace{-3pt}
\subsubsection{IDR Adaptation Module (IDR-AM)}
IDR-AM adopts a hierarchical adaptation process, consisting of a series of IDR Correction Blocks, IDR Adaptation Blocks and IDR Adaptation Groups.
Details are as follows.
\par
\noindent \textbf{1) IDR Correction Block (IDR-CB).}
The IDR-CB is designed to further optimize channel-wise IDR $R_{C}$ and spatial-wise IDR $R_{S}$.
The channel-level correction includes two FC layers with GELU \cite{gelu_32} and a residual connection, while the spatial-level correction includes two Conv layers with GELU and a residual connection. 
Sec.\ref{ablation} shows that this module can enhance the robustness of adaptation module to IDR estimation errors to a certain extent.
\par
\noindent\textbf{2) IDR Adaptation Block (IDR-AB).}
IDR-AB is the basic adaptation unit.
As shown in Fig.\ref{fig:overall}, 
taking LR features $f_{LR}^{in}$ and corrected IDRs as inputs, IDR-AB modulates IDRs from channel and spatial perspectives, and integrates IDRs into LR features.
Given its intricate design, we will elaborate on the internal structure of IDR-AB in detail in Sec.\ref{teacherstage} by combining the specific calculation process.
\par
\noindent \textbf{3) IDR Adaptation Group (IDR-AG).}
The IDR-AG is composed of eight IDR-ABs, followed by a ConvNeXt block \cite{convnext_24} and a 3$\times$3 Conv layer for further feature fusion. 
And a residual link is added between the input and output of an IDR-AG for restoring low-level details.
\par
Finally, the IDR-AM consists of eight IDR-AGs and a 3$\times$3 Conv layer.
A residual link is also applied to preserve the original image details and semantic information.
\par
\vspace{-2pt}
\vspace{-3pt}
\subsection{KD-based Model Learning}\label{SecTeKD}
\begin{figure*}[t] {\centering
      \centerline{\includegraphics[width=0.99\linewidth]{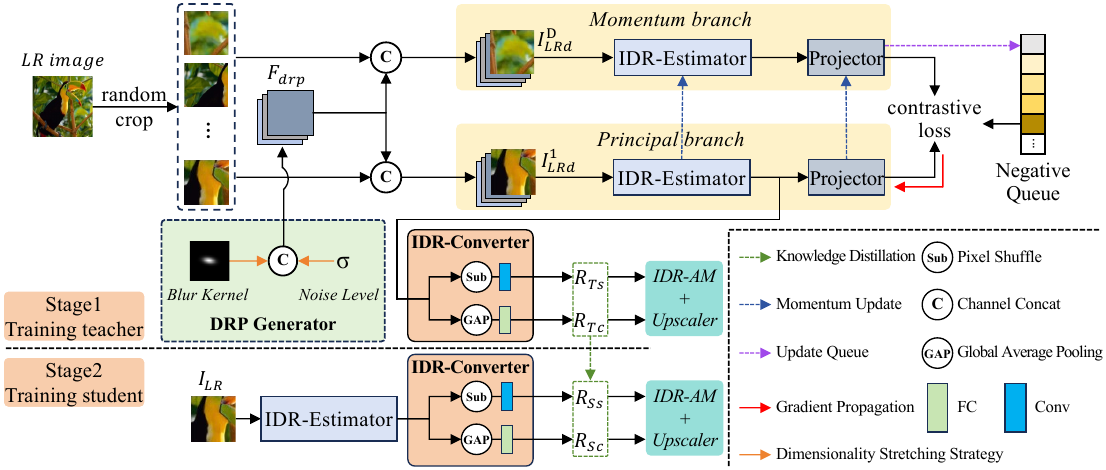}} \vspace{-4pt}
      \caption{
      The training illustration of our LightBSR, in which the teacher stage uses complex structures to enhance IDR discriminability.
      }
      \label{fig:contrast}
      }\vspace{-13pt}
\end{figure*}

\vspace{-3pt}
As shown in Fig.\ref{fig:contrast},
the training of proposed LightBSR is divided into two stages: 
For teacher training, contrastive learning (CL) and degradation reference prior (DRP) are introduced into the training process to reinforce the discriminability of the learned IDR latent space.
For student training, the degradation knowledge learned by the teacher is transferred to the student through feature alignment.
\vspace{-5pt} 
\subsubsection{Teacher Training}\label{teacherstage}
\vspace{-2pt} 
The CL-based teacher training framework mainly consists of a DRP generator, a principal branch, a momentum branch and a negative sample queue $\mathcal{Q}$.
Each branch includes an IDR-estimator and a projector (a two-layer FC network).
The training is divided into two stages, of which the first stage pre-trains the IDR-estimator by only updating two CL branches, and the second stage fine-tunes the whole model parameters.
Note that during teacher training, the IDR-estimator and the IDR-converter are separate.
\par
\noindent \textbf{1) Data Preparation.}
Given a batch of $B$ LR images, we randomly crop $D$ patches (size is $3 \times H \times W$) for each LR image to form $B$ positive sample sets.
For DRP generation, we refer to dimensionality stretching strategy \cite{srmd_20}.
Specifically, 
first, the blur kernel of size $k \times k$ is vectorized and projected into a $t$-dimensional space via PCA, resulting in a $t \times 1$ vector. 
Next, the noise level value $\sigma$ is replicated 3 times and concatenated with the blur kernel vector, yielding a $(t+3) \times 1$ vector. 
Then, this vector is stretched to match the LR patch dimensions, resulting in the tensor representation of DRP $F_{drp} \in\mathbb{R}^{(t+3) \times H \times W}$.
Finally, the input for teacher training is the concatenation of each patch and its corresponding $F_{drp}$, defined as $I_{LRd}\in\mathbb{R}^{B\times D \times (t+6) \times H \times W}$.
\par

\vspace{2pt}
\noindent \textbf{2) IDR-Estimator Pre-training.}
The first stage can be seen as pre-training the IDR-Estimator.
$I_{LRd}$ is fed simultaneously into the principal and momentum branches, producing $\mathcal{P}, \mathcal{M}\in\mathbb{R}^{B\times D \times 128}$, respectively. 
Then, the contrastive loss is computed based on $\mathcal{M}$, $\mathcal{P}$, and $\mathcal{Q}$, as follows:
\vspace{-0.2em}
\begin{small}
\begin{equation}
    L_{CL} = \frac{1}{D(D-1)}\sum\limits_{i=1}^D\sum\limits_{j=1}^D Sim(\mathcal{P}_i,\mathcal{M}_j), (i \neq j),
    \vspace{-0.6em}
\end{equation}
\end{small}
where the operator $Sim(\cdot)$ refers to \cite{infonce}, as follows:
\vspace{-0.2em}
\begin{small}
\begin{equation}
    \begin{split}
    & Sim(\mathcal{P}_i, \mathcal{M}_j) = \\ 
    &-\frac{1}{B} \sum_{k = 1}^{B}log\frac{exp(\mathcal{P}_i^k \cdot \mathcal{M}_j^k / \tau )}{ exp(\mathcal{P}_i^k \cdot \mathcal{M}_j^k / \tau )+ \sum_{l = 1}^{N}exp(\mathcal{P}_i^k \cdot \mathcal{Q}^{l} / \tau)},
    \end{split}
    \vspace{-0.6em}
\end{equation} 
\end{small}%
where $\tau$ denotes the temperature coefficient, and $N$ is the length of $\mathcal{Q}$.\ 
For the momentum branch, the parameters are updated using the momentum update strategy:
\vspace{-0.2em}
\begin{equation}
      \theta^\mathcal{M} \leftarrow \alpha \theta^\mathcal{M} + (1-\alpha)\theta^\mathcal{P},
\vspace{-0.2em}
\end{equation}
where $\theta^\mathcal{M}$ and $\theta^\mathcal{P}$ are the parameters of the momentum and principal branch, $\alpha\in\left[0,1\right]$ is the momentum coefficient.
\par
\noindent \textbf{3) Overall Model Fine tuning.}
The second stage expanded the parameter training scope based on the first stage.
Specifically, after obtaining the output of the IDR-estimator of the principal branch, it is further fed into the IDR-converter for spatial- and channel-domain transformation, 
outputting a spatial-wise IDR $R_{Ts}\in\mathbb{R}^{8 \times H \times W}$ and a channel-wise IDR $R_{Tc}\in\mathbb{R}^{48}$, respectively.
These two IDRs, along with the LR image features, are then fed into the IDR-AM and Upscaler to generate the HR image.
The L1 loss is used to measure the reconstruction error, as follows:
\vspace{-4pt}
\begin{equation}
      L_{SR}= \left\Vert {I}_{SR}-I_{HR}\right\Vert _{1},
      \vspace{-4pt}
\end{equation}
where ${I}_{SR}$ represents the output of the BSR network, $I_{HR}$ represents the HR version corresponding to the input LR image. 
The sum of $L_{SR}$ and $L_{CL}$ is defined as the total loss in the second stage for training the teacher network.
\par
\noindent \textbf{4) IDR Adaptation.}
As detailed in Sec.\ref{overviwer}, the adaptation process in IDR-AM is sequential, consisting of a series of IDR-AGs and a Conv layer, where each IDR-AG is composed of a series of IDR-ABs and two Conv blocks.
As shown in Fig.\ref{fig:overall}, 
taking output IDRs of IDR-EM and LR features as input, 
IDR-CB first optimizes IDRs to obtain $\widehat{R}_{C} $ and $ \widehat{R}_{S}$ and sends them to the corresponding IDR-AG and next IDR-CB. 
In IDR-AG, the IDR-AB receives optimized IDRs and LR features as inputs.
IDR-AB first divides the input LR features $f_{LR}^{in}$ into two parts along the channel dimension in a 1:3 ratio to fully use the redundancy between different channels and reduce computational complexity.
For spatial-wise modulation, $\frac{1}{4}f_{LR}^{in}$ is concatenated with $\widehat{R}_{S}$ along the channel dimension, followed by modulation through two 3$\times$3 Conv layers and a Sigmoid function. 
Next, the modulated spatial-wise IDR is multiplied with the original $\frac{1}{4}f_{LR}^{in}$, followed by a 5$\times$5 Conv layer for further fusion to obtain the spatial-wise degradation-adapted LR image features $f_{LR}^{S}$.
For channel-wise modulation, the remaining $\frac{3}{4}f_{LR}^{in}$ is first processed by a GAP operator, and then the pooled result is concatenated to $\widehat{R}_{C}$ along the channel dimension, followed by modulation through two FC layers and a Sigmoid function.
Next, the modulated channel-wise IDR is multiplied with the original $\frac{3}{4}f_{LR}^{in}$ to obtain the channel-wise degradation-adapted LR features $f_{LR}^{C}$.
Then $f_{LR}^{S}$ and $f_{LR}^{C}$ are concatenated and passed through a ConvNeXt block \cite{convnext_24} for information complementary processing.
Finally, a residual link is applied to integrate the original $f_{LR}^{in}$ into the processed result to enhance the image details, resulting in the final output $\hat{f}_{LR}^{in}$ of current IDR-AB.
After a series of hierarchical adaptations in IDR-AM, LR image features that fully integrate degradation information are obtained, denoted as $f_{LR}^{out}$, which would be used as input for the Upscaler.
\begin{table}[t]
     
      \renewcommand{\baselinestretch}{1.3}
      {\footnotesize\centerline{\tabcolsep=6.8pt
      \begin{tabular*}{0.48\textwidth}{c|cc|cccc}
      \bottomrule

    \multicolumn{1}{c|}{\multirow{1}[0]{*}{Method}} & \multicolumn{1}{c}{\multirow{1}[0]{*}{\shortstack{DRP}}} & \multicolumn{1}{c|}{\multirow{1}[0]{*}{\shortstack{CL}}} 
    & \multicolumn{1}{c}{\multirow{1}[0]{*}{Set5}} & \multicolumn{1}{c}{\multirow{1}[0]{*}{Set14}} & \multicolumn{1}{c}{\multirow{1}[0]{*}{B100}} & \multicolumn{1}{c}{\multirow{1}[0]{*}{Urban100}} \\
    \hline
    T1 & \multicolumn{1}{c}{} & \multicolumn{1}{c|}{} & 31.60  & 28.25  & 27.40 & 25.69 \\
    T2 & $\checkmark$     & \multicolumn{1}{c|}{} & 31.96 & 28.45 & 27.47 & 25.94 \\
    T3 & \multicolumn{1}{c}{} & $\checkmark$    & 31.75 & 28.34 & 27.43 & 25.77 \\
    T4(Ours) & $\checkmark$     & $\checkmark$     & \textbf{32.00} & \textbf{28.49} & \textbf{27.50} & \textbf{26.02} \\
      \toprule
\end{tabular*}}}
\vspace{-6pt}
 \caption{The effect of different training framework designs on the average PSNR under various kernel widths $\{1.2, 2.4, 3.6\}$.}
      \vspace{-8pt}
      \label{table:xiaorong1}
    
\end{table}
\begin{table}[t]
      
      \renewcommand{\baselinestretch}{1.3}
      {\footnotesize\centerline{\tabcolsep=4.0pt
      \begin{tabular*}{0.48\textwidth}{cccc|cccc}
      \bottomrule
     \multicolumn{1}{c|}{\multirow{1}[0]{*}{Method}} & \multicolumn{1}{c}{\multirow{1}[0]{*}{\shortstack{$\widehat{R}_{S}$}}} & \multicolumn{1}{c}{\multirow{1}[0]{*}{\shortstack{$\widehat{R}_{C}$}}} & \multicolumn{1}{c|}{\multirow{1}[0]{*}{IDR-CB}} & \multicolumn{1}{c}{\multirow{1}[0]{*}{Set5}} & \multicolumn{1}{c}{\multirow{1}[0]{*}{Set14}} & \multicolumn{1}{c}{\multirow{1}[0]{*}{B100}} & \multicolumn{1}{c}{\multirow{1}[0]{*}{Urban100}} \\
    \hline
    \multicolumn{1}{c|}{M1} &       &       &       & 29.25 & 27.01 & 26.48 & 24.21 \\
    \multicolumn{1}{c|}{M2} & \multicolumn{1}{c}{$\checkmark$} &       & \multicolumn{1}{c|}{$\checkmark$} & 31.86  & 28.43 & 27.46 & 25.87 \\
    \multicolumn{1}{c|}{M3} &       & \multicolumn{1}{c}{$\checkmark$} & \multicolumn{1}{c|}{$\checkmark$} & 31.92 & 28.44 & 27.47 & 25.91 \\
    \multicolumn{1}{c|}{M4(Ours)} & \multicolumn{1}{c}{$\checkmark$} & \multicolumn{1}{c}{$\checkmark$} & \multicolumn{1}{c|}{$\checkmark$} & \textbf{32.00} & \textbf{28.49} & \textbf{27.50} & \textbf{26.02} \\
    \multicolumn{1}{c|}{M5} & \multicolumn{1}{c}{$\checkmark$} & \multicolumn{1}{c}{$\checkmark$} &       & 31.94 & 28.42 & 27.47 & 25.93 \\
    \hline
    \multicolumn{1}{c|}{M4-} & \multicolumn{3}{c|}{M4 with error IDRs} & 28.15 & 25.95 & 25.79 & 23.37 \\
    \multicolumn{1}{c|}{M5-} & \multicolumn{3}{c|}{M5 with error IDRs} & 28.05 & 25.89 & 25.69 & 23.26 \\
    \toprule
\end{tabular*}}}
\vspace{-6pt}
\caption{The effect of different IDR-AM designs on the average PSNR under various kernel widths $\{1.2, 2.4, 3.6\}$.}
      \label{table:xiaorong2}
      \vspace{-8pt}
\end{table}
\noindent
\begin{figure}[t]
      \centering
      \includegraphics[width=0.99\linewidth]{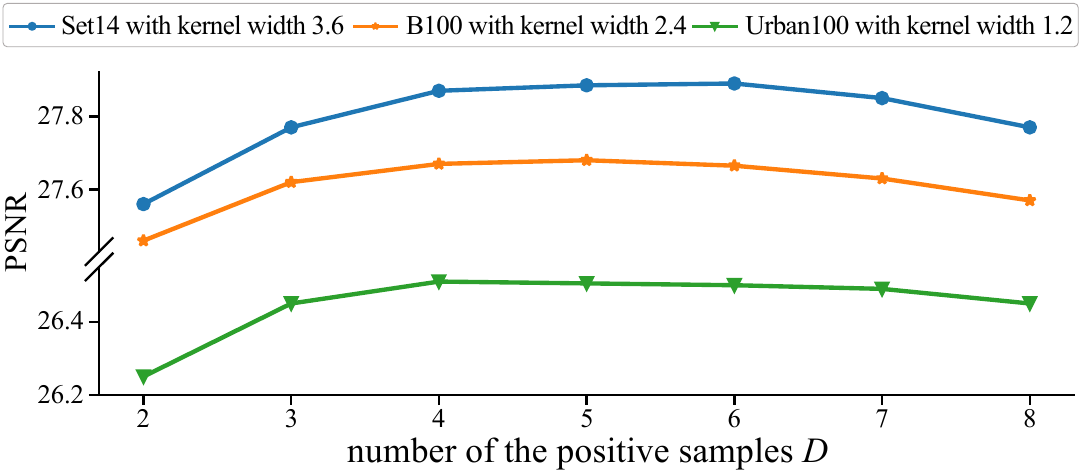}
      \vspace{-5pt}
      \caption{The effect of different numbers of positive samples.}
      \vspace{-18pt}
      \label{fig:positive_number}
\end{figure}
\vspace{-3pt}
\subsubsection{Knowledge Transfer}
\vspace{-2pt}
The student network mainly consists of an IDR-estimator IDR-E$_{S}$ and an IDR-converter IDR-C$_{S}$, with each component having the same structure as in the teacher network.
In terms of input, the student network only takes the LR image as input, fulfilling the inference requirements.
The spatial- and channel-wise IDRs output by IDR-C$_{S}$ are denoted as $R_{Ss}$ and $R_{Sc}$, respectively.
After training the teacher network, the IDR-estimator of the principal branch and IDR-converter are exported as the teacher, defined as IDR-E$_{T}$ and IDR-C$_{T}$, respectively, and the learned knowledge is transferred to the student network using the distillation learning technique.
Specifically, knowledge distillation is achieved by aligning the output between teacher and student converters,
and the training is also divided into two stages, with the first stage only updating IDR-E$_{S}$ and IDR-C$_{S}$, and the second stage optimizing the whole model parameters.
\par 
In the first stage, 
for spatial-domain distillation, $L_2$ loss is used to perform pixel-level knowledge transfer from $R_{Ts}$ to $R_{Ss}$, which can be expressed as: 
\vspace{-4pt}
\begin{equation}
      L_{2}=\left\Vert R_{Ts} - R_{Ss} \right\Vert _{2}.
\vspace{-4pt}
\end{equation}
\par
For channel-domain representations $R_{Tc}$ and $R_{Sc}$, the Kullback–Leibler (KL) divergence loss is employed to match their distributions and the L1 loss is used to minimize their absolute differences, as follows:
\begin{equation}
      L_{kl}=\sum\limits_{i=1}^{C} R_{Tc}^{norm}[i] log(\frac{R_{Tc}^{norm}[i]}{R_{Sc}^{norm}[i]}),
      \vspace{-1pt}
\end{equation}
\begin{equation}
      L_{1}=\left\Vert R_{Tc} - R_{Sc} \right\Vert _{1},
      \vspace{-1pt}
\end{equation}
where $C$ is the number of channels, and $R_{Tc}^{norm}$ and $R_{Sc}^{norm}$ are softmax-normalized outputs of $R_{Tc}$ and $R_{Sc}$, respectively. 
The total loss of knowledge distillation is:
\begin{equation}
      L_{DL}=L_{2}+L_{kl}+ \beta L_{1},
      \vspace{-3pt}
\end{equation}
where $\beta$ denotes the balance factor. 
In the second stage, $R_{Ss}$ and $R_{Sc}$ are further fed into IDR-AM and Upscaler for SR reconstruction, consistent with the teacher training. 
The sum of $L_{SR}$ and $L_{DL}$ is defined as the loss of the second stage for training the student network.
\vspace{-6pt}

\section{Experiments}
\vspace{-2pt}
\label{sec:experiment}

\subsection{Experimental Setup}\label{sec:s4.1}
\vspace{-3pt}
\noindent \textbf{Datasets preparation.} For training, we use 800 images from DIV2K \cite{div2k_33} and 2650 images from Flickr2K \cite{Flickr2k_34}, consistent with previous works \cite{dasr_5,kdsr}. 
For evaluation, we choose four standard benchmarks: Set5 \cite{set5_35}, Set14 \cite{set14_36}, B100 \cite{b100_37} and Urban100 \cite{urban100_38}. 
The LR-HR sample pairs required for training follow the classical degradation model:\par
\vspace{-7pt}
\begin{equation}
      I_{LR}=(I_{HR} \otimes k)\downarrow _s + n \label{eq:fzth1},
      \vspace{-1pt}
\end{equation}
where $I_{HR}$ is the original HR image, $I_{LR}$ is the degraded LR image, $\otimes$ denotes convolution operation, $k$ denotes the blur kernel, $\downarrow _s$ denotes the bicubic downsampling with scale factor $s$, and $n$ refers to additive Gaussian white noise. 
In experiments, two degradation settings are employed:
\par
\noindent \textbf{Setting 1:} only contains isotropic Gaussian blur kernels. The kernel size is fixed to 21 $\times$ 21, and the range of kernel width is set to [0.2, 4.0] for $\times$4 SR.\par
\noindent \textbf{Setting 2:} considers anisotropic Gaussian blur kernels and Gaussian white noise for $\times$4 SR. The anisotropic Gaussian blur kernel characterized by a Gaussian probability density function ${N}$(0, $\Sigma$), where covariance matrix $\Sigma$ is determined by two eigenvalues $\lambda_1$, $\lambda_2$ $\sim$ $\mathsf{U}$(0.2, 4) and a rotation angle $\theta$ $\sim$ $\mathsf{U}$(0, $\pi$). The kernel size is fixed to 21 $\times$ 21 and the range of Gaussian white noise is set to [0, 25]. 

\begin{figure*}[t] {\centering
      \centerline{\includegraphics[width=0.97\textwidth]{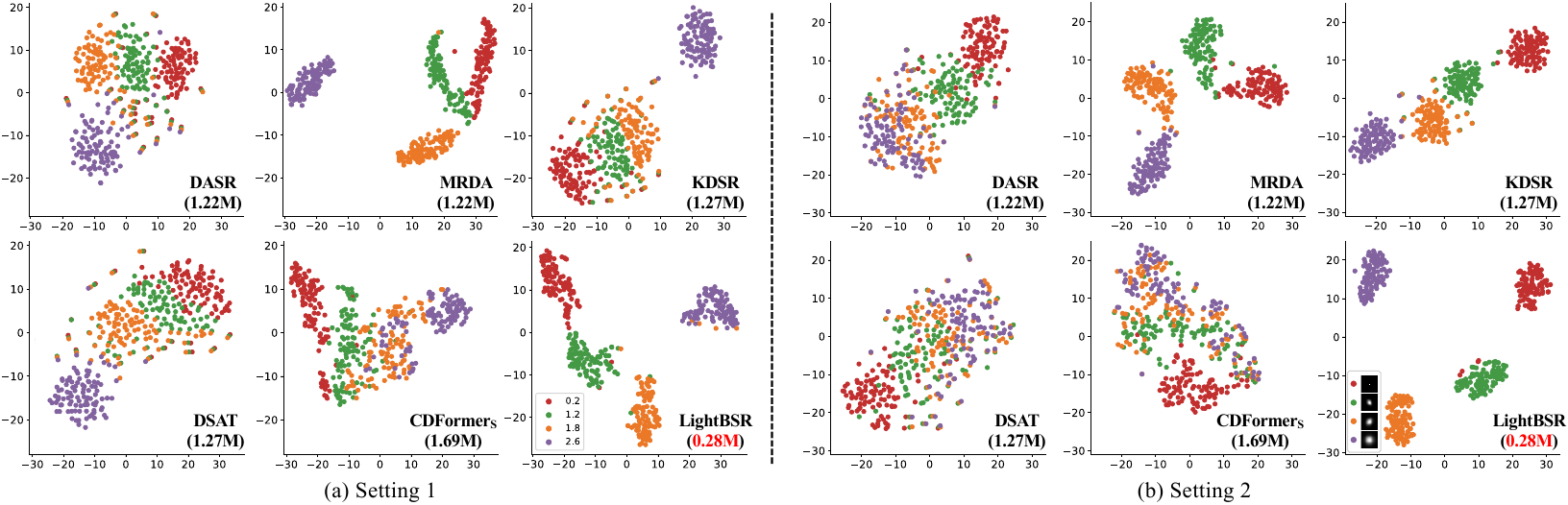}}\vspace{-4pt}
      \caption{The t-SNE \cite{tsne_53} plots of IDR distributions on the B100 \cite{b100_37} benchmark. 
      a) 4 different isotropic blur kernels are selected under \textit{Degradation Setting 1}. 
      b) 4 different anisotropic blur kernels are chosen under \textit{Degradation Setting 2}, with the noise level set to 4.}
      \label{fig:exper_tsne}
      \vspace{-8pt}
      }
\end{figure*}
\begin{table*}[t]
     
      \renewcommand{\baselinestretch}{1.0}
      {\footnotesize\centerline{\tabcolsep=4.2pt
      \begin{tabular*}{\textwidth}{cccccccccccccccc}
      \midrule
      \multicolumn{1}{c}{\multirow{2}[1]{*}{\textbf{Method}}} & \multicolumn{1}{c}{\multirow{2}[1]{*}{\textbf{Param.}}} & \multicolumn{1}{c}{\multirow{2}[1]{*}{\textbf{GFLOPs}}} &  \textbf{Dataset} & \multicolumn{3}{c}{\textbf{Set5}} & \multicolumn{3}{c}{\textbf{Set14}} & \multicolumn{3}{c}{\textbf{B100}} & \multicolumn{3}{c}{\textbf{Urban100}} \\
         \cmidrule(r){4-4}  \cmidrule(r){5-7}  \cmidrule(r){8-10} \cmidrule(r){11-13} \cmidrule(r){14-16}   &   &    &  \textbf{Kernel width} & \textbf{1.2} & \textbf{2.4} & \textbf{3.6} & \textbf{1.2} & \textbf{2.4} & \textbf{3.6} & \textbf{1.2} & \textbf{2.4} & \textbf{3.6} & \textbf{1.2} & \textbf{2.4} & \textbf{3.6} \\
          \midrule
          \multicolumn{1}{c}{\textbf{Bicubic}} & \multicolumn{1}{c}{---} &\multicolumn{1}{c}{---} & \multicolumn{1}{c}{---}  & 27.69 & 25.99 & 24.45 & 25.60  & 24.39 & 23.25 & 25.58 & 24.67 & 23.80  & 22.73 & 21.76 & 20.84 \\
         \midrule
        \multicolumn{1}{c}{\textbf{IKC}} & \multicolumn{1}{c}{5.3M} &\multicolumn{1}{c}{2876.30} & \multicolumn{1}{c}{CNN} & 31.76 & 30.35 & 30.26 & 28.44 & 28.17 & 26.63 & 27.43 & 27.28 & 26.41 & 25.63 & 25.02 & 24.07 \\
        \multicolumn{1}{c}{\textbf{DAN}} & \multicolumn{1}{c}{4.3M} & \multicolumn{1}{c}{1250.38} & \multicolumn{1}{c}{CNN} & 32.22 & 31.96 & 30.94 & 28.65 & 28.54 & 27.68 & 27.65 & 27.58 & 26.95 & 26.20  & 25.96 & 25.08 \\
        \multicolumn{1}{c}{\textbf{DCLS}} & \multicolumn{1}{c}{13.6M} & \multicolumn{1}{c}{496.72} & \multicolumn{1}{c}{CNN} & 32.35 & \textcolor[rgb]{ 1,  0,  0}{\textbf{32.19}} & \textcolor[rgb]{ 1,  0,  0}{\textbf{31.14}} & 28.66 & 28.57 & 27.78 & 27.73 & 27.65 & \textcolor[rgb]{ 0,  .439,  .753}{\underline{27.02}}  & \textcolor[rgb]{ 1,  0,  0}{\textbf{26.50}} & \textcolor[rgb]{ 1,  0,  0}{\textbf{26.24}} & \textcolor[rgb]{ 1,  0,  0}{\textbf{25.34}}  \\
        \midrule
        \multicolumn{1}{c}{\textbf{DASR}} & \multicolumn{1}{c}{5.8M} & \multicolumn{1}{c}{208.80} & \multicolumn{1}{c}{CNN} & 31.92 & 31.75 & 30.59 & 28.44 & 28.28 & 27.45 & 27.52 & 27.43 & 26.83 & 25.69 & 25.44 & 24.66 \\
        \multicolumn{1}{c}{\textbf{IDMBSR}} & \multicolumn{1}{c}{4.2M} & \multicolumn{1}{c}{---} & \multicolumn{1}{c}{CNN} & 31.90  & 31.78 & 30.68 & 28.50  & 28.36 & 27.60  & 27.58 & 27.51 & 26.90  & 25.91 & 25.68 & 24.91 \\
        \multicolumn{1}{c}{\textbf{MRDA}} & \multicolumn{1}{c}{5.8M} & \multicolumn{1}{c}{386.25} & \multicolumn{1}{c}{CNN} & 32.36 & 32.11 & 30.89  & 28.67 & 28.57 & 27.62 & 27.67 & 27.58 & 26.91 & 26.26 & 26.02 & 25.08 \\
        \multicolumn{1}{c}{\textbf{KDSR}} & \multicolumn{1}{c}{5.8M} & \multicolumn{1}{c}{258.14} & \multicolumn{1}{c}{CNN} & 32.34 & 32.13 & 31.02  & 28.66 & 28.55 & \textcolor[rgb]{ 1,  0,  0}{\textbf{27.81}} & 27.67 & 27.59 & 26.97 & 26.29 & 26.05 & 25.20  \\
         \multicolumn{1}{c}{\textbf{DSAT}} & \multicolumn{1}{c}{15.6M} & \multicolumn{1}{c}{895.03} & \multicolumn{1}{c}{Transformer} & \textcolor[rgb]{ 1,  0,  0}{\textbf{32.51}} & 32.00 & 30.31 & \textcolor[rgb]{ 0,  .439,  .753}{\underline{28.77}} & 28.50 & 27.51 & \textcolor[rgb]{ 0,  .439,  .753}{\underline{27.76}} & \textcolor[rgb]{ 0,  .439,  .753}{\underline{27.66}} & \textcolor[rgb]{ 0,  .439,  .753}{\underline{27.02}} & \textcolor[rgb]{ 0,  .439,  .753}{\underline{26.43}} & 25.95 & 24.89 \\
        \multicolumn{1}{c}{\textbf{CDFormer\textsubscript{S}}} & \multicolumn{1}{c}{11.9M} & \multicolumn{1}{c}{631.53} & \multicolumn{1}{c}{Transformer} & 32.36 & 32.14 & \textcolor[rgb]{ 0,  .439,  .753}{\underline{31.07}} & \underline{\textcolor[rgb]{ 0,  .439,  .753}{28.77}} & \underline{\textcolor[rgb]{ 0,  .439,  .753}{28.60}} & 27.67 & 27.74 & 27.64 & 27.00 & 26.39 & 26.09 & 25.19 \\
        \midrule
        \multicolumn{1}{c}{\textbf{LightBSR}} & \multicolumn{1}{c}{\textcolor[rgb]{ 1,  0,  0}{\textbf{3.1M}}} & \multicolumn{1}{c}{\textcolor[rgb]{ 1,  0,  0}{\textbf{206.16}}} & \multicolumn{1}{c}{CNN} & \textcolor[rgb]{ 0,  .439,  .753}{\underline{32.47}} & \textcolor[rgb]{ 0,  .439,  .753}{\underline{32.17}} & 31.06 & \textcolor[rgb]{ 1,  0,  0}{\textbf{28.82}} & \textcolor[rgb]{ 1,  0,  0}{\textbf{28.65}} & \textcolor[rgb]{ 0,  .439,  .753}{\underline{27.79}} & \textcolor[rgb]{ 1,  0,  0}{\textbf{27.77}} & \textcolor[rgb]{ 1,  0,  0}{\textbf{27.69}} & \textcolor[rgb]{ 1,  0,  0}{\textbf{27.04}} & \textcolor[rgb]{ 1,  0,  0}{\textbf{26.50}} & \textcolor[rgb]{ 0,  .439,  .753}{\underline{26.20}} & \textcolor[rgb]{ 0,  .439,  .753}{\underline{25.31}}\\
          \toprule
\end{tabular*}}}
\vspace{-6pt}
 \caption{Quantitative comparison of PSNR metric for different methods under \textit{Degradation Setting 1} on $\times$4 SR. 
      The best and second-best results are highlighted in \textcolor[rgb]{ 1,  0,  0}{\textbf{red}} and \textcolor[rgb]{ 0,  .439,  .753}{\underline{blue}}, respectively. GFLOPs are calculated based on a 256 $\times$ 256 image size.}
      \label{table:x4_iso}
      \vspace{-14pt}
\end{table*}

\noindent \textbf{Implementation Details.} For preprocessing, HR images are augmented via random rotation and flipping and the LR patch size is set to 64 $\times$ 64.
For training, the batchsize $B$ is set to 64 and the number of the positive sample $D$ is set to 4. 
During IDR modeling, 
the PCA dimension $t$ is set to 15, the temperature coefficient $\tau$ is set to 0.07, the momentum coefficient $\alpha$ is set to 0.999, and the balance coefficient $\beta$ in $L_{DL}$ is set to 0.1. 
For both teacher and student training, 
in the first stage, learning rate is set to 2e-4 for 100 epochs. 
In the second stage, cosine annealing strategy \cite{cosine} is used to gradually decrease the learning rate from 2e-4 to 1e-6 over 600 epochs. 
Adam optimizer \cite{adam} is used for training.

\begin{figure*}[t] 
      {
      \centering
    \centerline{\includegraphics[width=0.96\linewidth]{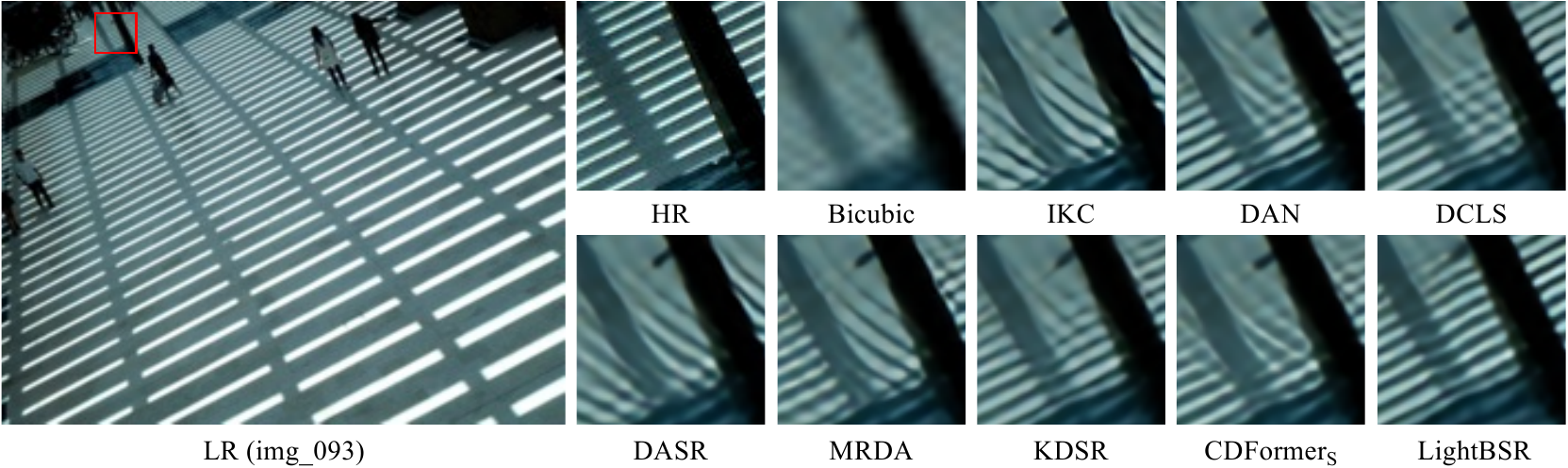}}
    \vspace{-5pt}
      \caption{$\times$4 SR results for isotropic Gaussian blur kernel width 1.2 on the ``img\_093'' of Urban100 benchmark.}
      \label{fig:x4_iso}
      }
      \vspace{-5pt}
\end{figure*}
\begin{table*}[t]
     
      \renewcommand{\baselinestretch}{1.1}
      {\footnotesize\centerline{\tabcolsep=2.5pt\begin{tabular*}{\textwidth}{c|cccccccccccccc}
        \toprule
        \multicolumn{1}{c}{\multirow{2}[-2]{*}{\textbf{Noise}}} & \multicolumn{1}{c}{\multirow{2}[-2]{*}{\textbf{Method}}} & \multicolumn{1}{c}{\multirow{2}[-2]{*}{\textbf{Param.}}} & 
     \begin{minipage}[b]{0.15\columnwidth}
      \centering
      \raisebox{-.45\height}{\includegraphics[width=0.45\linewidth]{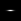}} 
      \end{minipage} & 
      \textbf{\begin{minipage}[b]{0.15\columnwidth}
      \centering
      \raisebox{-.45\height}{\includegraphics[width=0.45\linewidth]{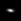}} 
      \end{minipage}} & 
      \textbf{\begin{minipage}[b]{0.15\columnwidth}
            \centering
            \raisebox{-.45\height}{\includegraphics[width=0.45\linewidth]{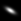}} 
            \end{minipage}} & 
      \textbf{\begin{minipage}[b]{0.15\columnwidth}
      \centering
      \raisebox{-.45\height}{\includegraphics[width=0.45\linewidth]{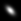}} 
      \end{minipage}} & 
      \textbf{\begin{minipage}[b]{0.15\columnwidth}
            \centering
            \raisebox{-.45\height}{\includegraphics[width=0.45\linewidth]{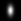}} 
      \end{minipage}} & 
      \textbf{\begin{minipage}[b]{0.15\columnwidth}
            \centering
            \raisebox{-.45\height}{\includegraphics[width=0.45\linewidth]{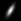}} 
      \end{minipage}} & 
      \textbf{\begin{minipage}[b]{0.15\columnwidth}
            \centering
            \raisebox{-.45\height}{\includegraphics[width=0.45\linewidth]{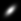}} 
      \end{minipage}} & 
      \textbf{\begin{minipage}[b]{0.15\columnwidth}
            \centering
            \raisebox{-.45\height}{\includegraphics[width=0.45\linewidth]{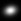}} 
      \end{minipage}} & 
      \begin{minipage}[b]{0.15\columnwidth}
            \centering
            \raisebox{-.45\height}{\includegraphics[width=0.45\linewidth]{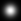}} 
      \end{minipage} \\ [0.28cm]
      \hline
       \multicolumn{1}{c|}{\multirow{6}[6]{*}{5}} 
          & \textbf{DnCNN+DAN} & 650K+4.33M & 26.18 & 26.14  & 24.89 & 24.64  & 24.61 & 24.52  & 24.37 & 24.01  & 23.63 \\
          & \textbf{DnCNN+DCLS} & 650K+13.63M & 26.15  & 26.12 & 24.89 & 24.66 & 24.62 & 24.52 & 24.39 & 24.03 & 23.65 \\
          & \textbf{DASR} & 5.84M & 26.60  & 26.52 & 25.87 & 25.69 & 25.64 & 25.60  & 25.48 & 25.23 & 24.85 \\
          & \textbf{MRDA} & 5.84M & 26.73 & 26.68 & 25.98 & 25.80  & 25.77 & 25.71 & 25.59 & 25.29 & 24.82 \\
          & \textbf{KDSR} & 5.80M & 26.73 & \underline{\textcolor[rgb]{ 0,  .439,  .753}{26.70}} & \underline{\textcolor[rgb]{ 0,  .439,  .753}{26.02}} & \underline{\textcolor[rgb]{ 0,  .439,  .753}{25.87}} & \underline{\textcolor[rgb]{ 0,  .439,  .753}{25.83}} & \underline{\textcolor[rgb]{ 0,  .439,  .753}{25.76}} & \underline{\textcolor[rgb]{ 0,  .439,  .753}{25.65}} & \underline{\textcolor[rgb]{ 0,  .439,  .753}{25.37}} & 24.94 \\
          & \textbf{DSAT} & 15.64M & 26.71 & 26.66 & 25.95 & 25.80  & 25.74 & 25.66 & 25.55 & 25.30 & 24.85 \\
           & \textbf{CDFormer\textsubscript{S}} & 11.86M & \underline{\textcolor[rgb]{ 0,  .439,  .753}{26.77}} & 26.69 & 26.00 & 25.85 & 25.82 & 25.75 & 25.64 & 25.34  & \underline{\textcolor[rgb]{ 0,  .439,  .753}{24.97}} \\
          & \textbf{LightBSR} & \textcolor[rgb]{ 1,  0,  0}{\textbf{3.09M}} & \textcolor[rgb]{ 1,  0,  0}{\textbf{26.82}} & \textcolor[rgb]{ 1,  0,  0}{\textbf{26.76}} & \textcolor[rgb]{ 1,  0,  0}{\textbf{26.06}} & \textcolor[rgb]{ 1,  0,  0}{\textbf{25.91}} & \textcolor[rgb]{ 1,  0,  0}{\textbf{25.87}} & \textcolor[rgb]{ 1,  0,  0}{\textbf{25.81}} & \textcolor[rgb]{ 1,  0,  0}{\textbf{25.70}} & \textcolor[rgb]{ 1,  0,  0}{\textbf{25.41}} & \textcolor[rgb]{ 1,  0,  0}{\textbf{25.05}} \\
    \hline
    \multicolumn{1}{c|}{\multirow{6}[5]{*}{10}}
          & \textbf{DnCNN+DAN} & 650K+4.33M & 25.69  & 25.63  & 24.62 & 24.42  & 24.39 & 24.29  & 24.18 & 23.89  & 23.54 \\
          & \textbf{DnCNN+DCLS} & 650K+13.63M & 25.67 & 25.62 & 24.62 & 24.43 & 24.40 & 24.29 & 24.19 & 23.91 & 23.56 \\
          & \textbf{DASR} & 5.84M & 26.02 & 25.94 & 25.27 & 25.11 & 25.08 & 25.03 & 24.93 & 24.69  & 24.33 \\
          & \textbf{MRDA} & 5.84M & 26.12 & 26.05 & 25.36 & 25.20  & 25.18 & 25.12 & 25.01 & 24.76 & 24.36 \\
          & \textbf{KDSR} & 5.80M & 26.11  & 26.05 & \underline{\textcolor[rgb]{ 0,  .439,  .753}{25.38}} & \underline{\textcolor[rgb]{ 0,  .439,  .753}{25.24}} & \underline{\textcolor[rgb]{ 0,  .439,  .753}{25.21}} & \underline{\textcolor[rgb]{ 0,  .439,  .753}{25.15}} & \underline{\textcolor[rgb]{ 0,  .439,  .753}{25.05}} & \underline{\textcolor[rgb]{ 0,  .439,  .753}{24.81}} & \underline{\textcolor[rgb]{ 0,  .439,  .753}{24.46}} \\
          & \textbf{DSAT} & 15.64M & 26.07 & 26.04 & 25.29 & 25.18  & 25.14 & 25.09 & 24.97 & 24.72 & 24.36 \\
           & \textbf{CDFormer\textsubscript{S}} & 11.86M & \underline{\textcolor[rgb]{ 0,  .439,  .753}{26.14}} & \underline{\textcolor[rgb]{ 0,  .439,  .753}{26.06}} & 25.37 & 25.22 & 25.20 & \underline{\textcolor[rgb]{ 0,  .439,  .753}{25.15}} & 25.04 & 24.78  & 24.45 \\
          & \textbf{LightBSR} & \textcolor[rgb]{ 1,  0,  0}{\textbf{3.09M}} & \textcolor[rgb]{ 1,  0,  0}{\textbf{26.18}} & \textcolor[rgb]{ 1,  0,  0}{\textbf{26.10}} & \textcolor[rgb]{ 1,  0,  0}{\textbf{25.41}} & \textcolor[rgb]{ 1,  0,  0}{\textbf{25.27}} & \textcolor[rgb]{ 1,  0,  0}{\textbf{25.24}} & \textcolor[rgb]{ 1,  0,  0}{\textbf{25.19}} & \textcolor[rgb]{ 1,  0,  0}{\textbf{25.08}} & \textcolor[rgb]{ 1,  0,  0}{\textbf{24.83}} & \textcolor[rgb]{ 1,  0,  0}{\textbf{24.53}} \\

      \toprule
\end{tabular*}}}
\vspace{-8pt}
 \caption{Quantitative comparison of PSNR metric for different methods on the B100 benchmark under \textit{Degradation Setting 2} on $\times$4 SR. The best and second-best results are highlighted in \textcolor[rgb]{ 1,  0,  0}{\textbf{red}} and \textcolor[rgb]{ 0,  .439,  .753}{\underline{blue}}, respectively.}\label{table:x4_aniso}
 \vspace{-16pt}
\end{table*}

\vspace{-5pt}
\subsection{Ablation Study}\label{ablation}
\vspace{-2pt}
We conduct a detailed analysis of the effects of each component in the proposed method under \textit{Degradation Setting 1}. Additional experimental results under \textit{Degradation Setting 2} are provided in the supplementary material.
\vspace{-6pt}
\subsubsection{Core Components for Training}
\noindent \textbf{1) Effect of DRP and CL.} 
These two are used to enhance the discriminability of IDRs. The ablation results are shown in Table \ref{table:xiaorong1}.
Using naive model T1 without any DRP or CL as the baseline, adding DRP (T2) or introducing CL (T3) both effectively improve performance, with average Peak signal-to-noise ratio (PSNR) on four benchmarks exceeding that of T1 by 0.22 dB and 0.09 dB, respectively.
Moreover, the combination of DRP and CL, i.e. T4 (ours), can further improve performance, demonstrating the effectiveness and compatibility of the two components.
\par
\noindent \textbf{2) Impact of Positive Sample Number.}
To verify the impact of the number of positive samples, we gradually increase the value of $D$ from 2 to 8, with the results shown in Fig.\ref{fig:positive_number}. 
When $D$ increased from 2 to 4, the performance on various benchmarks improves significantly, indicating that more positive samples can provide additional beneficial reference prior. 
However, when $D$ exceeds 4, the performance reaches saturation or even declines, possibly due to excessive redundant information impairing the estimator’s robustness against degradation distribution shifts \cite{redundant}. 
\vspace{-2pt}

\begin{figure*}[t] 
      {
      \centerline{\includegraphics[width=0.99\linewidth]{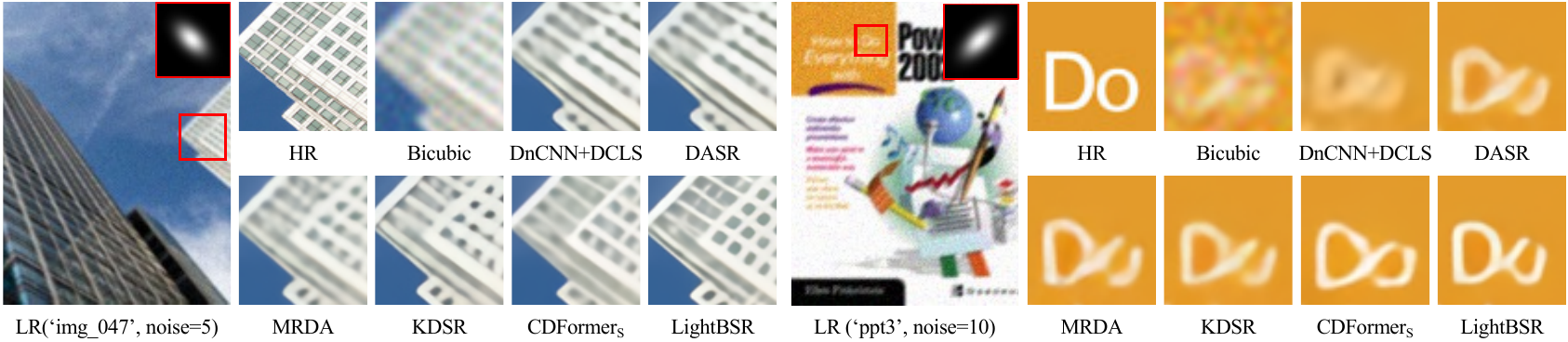}} 
      \vspace{-6pt}
      \caption{$\times$4 SR results for different anisotropic blur kernels and noise levels on the ``img\_047'' of Urban100 and the ``ppt3'' of Set14.}
      \vspace{-15pt}
      \label{fig:x4_aniso_5_10}
      }
\end{figure*}
\begin{figure}[t]
      \centering
      \includegraphics[width=\linewidth]{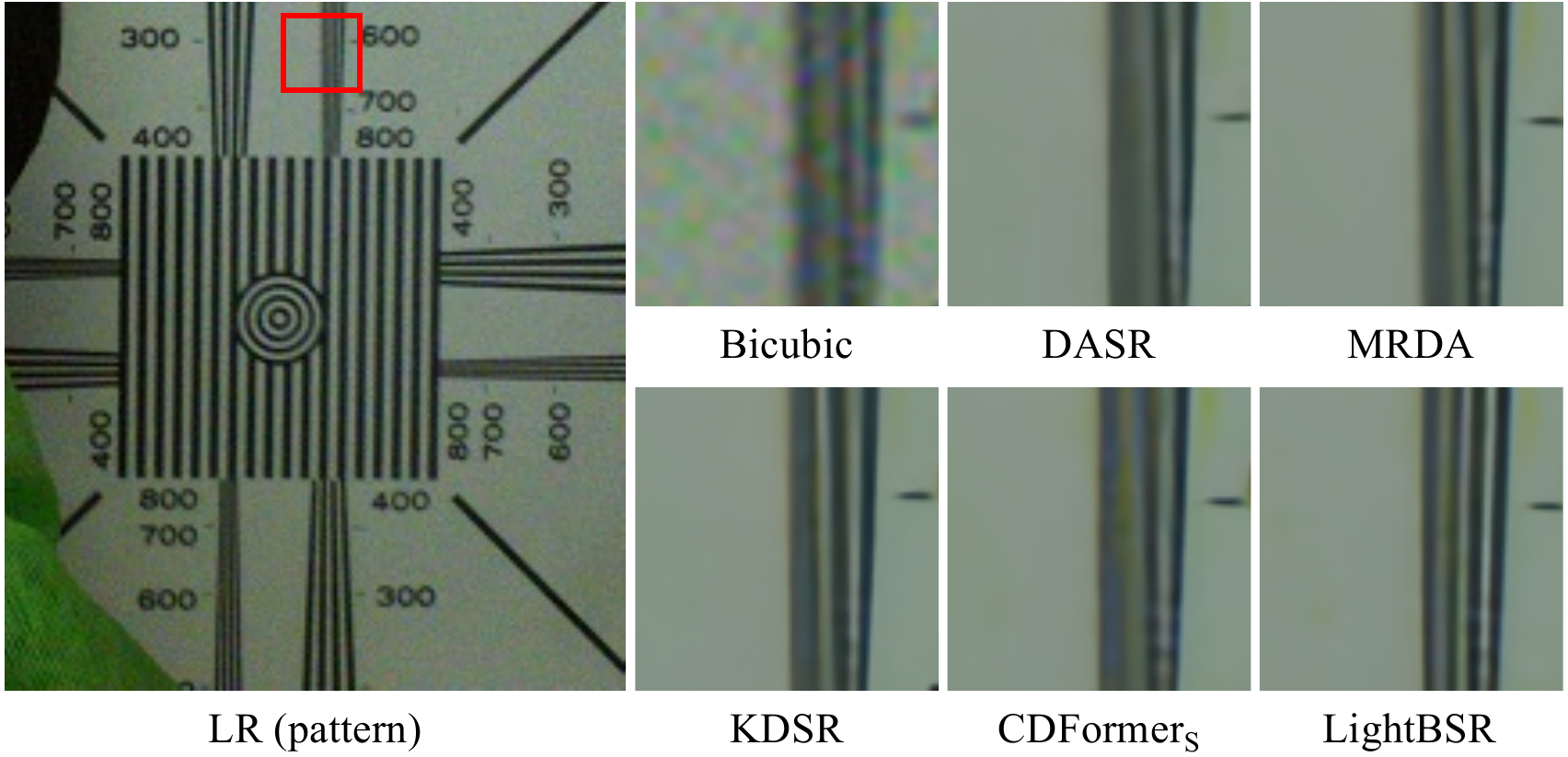}
      \vspace{-17pt}
      \caption{$\times$4 SR results on the ``pattern'' image of the RealWorld38 for various methods under \textit{Degradation Setting 2}.}
      \vspace{-18pt}
      \label{fig:real}
\end{figure}

\subsubsection{Core Components of IDR-AM}
\vspace{-2pt}
We perform detailed ablation studies on the core components of IDR-AM, namely IDR-AB and IDR-CB, and the ablation results are shown in Table \ref{table:xiaorong2}. 

\noindent \textbf{1) Effect of IDR-AB.}
There are two degradation modulation branches in IDR-AB, corresponding to channel and spatial perspectives. 
Compared to the baseline (M1) without any modulation, employing only spatial-domain modulation (M2) or channel-domain modulation (M3) both improves SR performance, with the average PSNR increasing by 1.67 dB and 1.7 dB, respectively, validating the necessity of modulation from both perspectives. 
And the combination of the two branches, i.e. M4 (ours), further improves the PSNR, showing the complementarity of the two.
\par
\noindent \textbf{2) Effect of IDR-CB.}
The variant M5 without IDR-CB shows a 0.06 dB decrease in average PSNR compared to M4 (ours), preliminarily showing the effectiveness of this component.
To further verify the effect of IDR-CB, we add random values ranging from 0 to 1 to the output of the IDR-Estimator, simulating incorrect IDRs and the results are shown at the bottom of Table \ref{table:xiaorong2}. 
It can be seen that when receiving error IDRs, M5- (without IDR-CB) has an average PSNR 0.09 dB lower than M4- (with IDR-CB), showing that IDR-CB can improve the robustness of SR network to error degradation estimation to a certain extent.\par
\vspace{-3pt}
\subsection{Comparison with Previous Methods}
\vspace{-2pt}
We compare our approach under two degradation settings against the following BSR methods: IKC \cite{ikc_14}, DAN \cite{dan_13}, DCLS \cite{DCLS}, DASR \cite{dasr_5}, IDMBSR \cite{idmbsr_7}, MRDA \cite{mrda_9}, KDSR \cite{kdsr}, DSAT \cite{DSAT} and CDFormer\textsubscript{S} \cite{CDFormer}.
Among these methods, IKC \cite{ikc_14}, DAN \cite{dan_13} and DCLS \cite{DCLS} follow EDE-BSR paradigm, while the others follow IDE-BSR paradigm.
The out-of-distribution (OOD) degradation evaluations are provided in the supplementary material.
\par
\vspace{-3pt}
\subsubsection{Experiments in Degradation Setting 1}
\vspace{-2pt}
\noindent \textbf{Quality of IDRs.}
We apply four different blur kernel widths to generate LR images for the B100 benchmark \cite{b100_37}, feeding them into the degradation estimators of DASR \cite{dasr_5}, MRDA \cite{mrda_9}, KDSR \cite{kdsr}, DSAT \cite{DSAT}, CDFormer\textsubscript{S} \cite{CDFormer} and LightBSR. 
The t-SNE plots of IDRs output by each method are shown in Fig.\ref{fig:exper_tsne}(a). 
Our method achieves better clustering of the same degradation and separation of different degradations, 
which shows the superiority of our method in discriminative implicit degradation estimation.
\par
\noindent \textbf{Quantitative evaluation.}
We use three different blur kernel widths \{1.2, 2.4, 3.6\} to evaluate the SR performance of various methods across four classic benchmarks. 
The comparisons in Table \ref{table:x4_iso} show that:
(1) Compared to the best EDE-based BSR method \cite{DCLS}, 
our approach achieves superior or matching performance under 8 out of 12 degradation conditions, 
while reducing the number of parameters by 78\% and computation by 59\%, 
showing the great potential of the IDE-based BSR paradigm.
(2) Compared with existing IDE-BSR methods, LightBSR achieves SOTA performance with fewer parameters and computation, showing the superiority of our method. 
For example, compared to CDFormer\textsubscript{S} \cite{CDFormer}, our method, which is based on CNN architecture, achieves superior results with only 26\% of the parameters and 32\% of the computational cost used by CDFormer\textsubscript{S}, demonstrating that CNNs remain highly competitive in the BSR task. 
\par
\noindent \textbf{Qualitative evaluation.}
The SR results of different methods under \textit{Degradation Setting 1} are shown in Fig.\ref{fig:x4_iso}. 
Our method recovers the overall structure of the image more comprehensively and restore complex architectural texture details more clearly and completely. 
This shows that the discriminability of IDRs is crucial for image restoration. 
\vspace{-4pt}
\subsubsection{Experiments in Degradation Setting 2}
\noindent \textbf{Quality of IDRs.}
We use four different anisotropic blur kernels and set the noise level to 4 to generate LR images for the B100 benchmark \cite{b100_37}. 
The IDRs generation and t-SNE visualization follow \textit{Degradation Setting 1}, and the results are shown in Fig.\ref{fig:exper_tsne}(b).
We can see that LightBSR also achieves the best discriminative IDR modeling. 
\par
\noindent \textbf{Quantitative evaluation.}
Here we select nine different anisotropic blur kernels and two noise levels to evaluate various methods on B100 benckmark \cite{b100_37}, with results shown in Table \ref{table:x4_aniso}. 
It is evident that our LightBSR achieves SOTA performance under all combinations of blur kernels and noise levels. 
Compared to the EDE-based DCLS \cite{DCLS}, LightBSR exceeds its average PSNR under all degradation conditions by 0.97 dB, corroborating the superiority of the IDE strategy in complex BSR tasks with noise interference. 
Compared to existing IDE-based methods, our method also improves performance with fewer parameters. 
For instance, LightBSR outperforms CDFormer\textsubscript{S} \cite{CDFormer} by 0.05 dB in average PSNR across all blur kernel and noise combinations, 
while the number of parameters is only 26\% of CDFormer\textsubscript{S}.
\par
\noindent \textbf{Qualitative evaluation.}
The SR results of different methods under \textit{Degradation Setting 2} are shown in Fig.\ref{fig:x4_aniso_5_10}. 
Our LightBSR is able to restore rich details clearly and accurately, even under severe degradation conditions.
In contrast, the SR images generated by DASR \cite{dasr_5}, MRDA \cite{mrda_9}, KDSR \cite{kdsr} and CDFormer\textsubscript{S} \cite{CDFormer} are either highly blurred or difficult to recover some complex texture details.
\vspace{-4pt}
\subsubsection{Experiments in Real Degradation Scenario}
\vspace{-1pt}
We use the “pattern” image from the RealWorld38 \cite{dasrReal} to evaluate the generalizability of all methods to complex, unknown real degradation.
Since no corresponding HR images are available, we use directly various models trained with \textit{Degradation Setting 2} to perform qualitative comparisons.
As shown in Fig.\ref{fig:real}, even if high-frequency details are lost, our method can still clearly restore line textures. More comparisons are provided in the supplementary material.
\vspace{-6pt}


\section{Conclusion}
\vspace{-3pt}
\label{sec:conclusion}
In this work, we demonstrate the importance of discriminative IDRs for the BSR task and reveal a novel lightweight design idea of BSR model.
The proposed LightBSR follows the KD-based training framework, in which the teacher stage uses a well-designed degradation-prior-constrained contrastive learning technique to enhance the discriminability of the learned IDR latent space, and the student stage uses a feature alignment technique to transfer the learned degradation-related knowledge to the structurally simple student for practical reasoning.
Extensive experiments show the effectiveness of our method, achieving excellent quantitative and qualitative results with minimal parameters and computation.
Additionally, our work also shows that the CNN architecture still has great potential in the blind SR task.

\par

\noindent \textbf{Acknowledgement.} This work was supported by the Natural Science Foundation of China (No. 62192782, 62036011), Beijing Natural Science Foundation (L223003), the Project of Beijing Science and Technology Committee (No. Z231100005923046).

{
    \small
    \bibliographystyle{ieeenat_fullname}
    \bibliography{main}
}

\clearpage
\setcounter{page}{1}
\maketitlesupplementary

Sec.\ref{add_ablation} presents the ablation study of our method under \textit{Degradation Setting 2}. 
Sec.\ref{unseen} provides implicit degradation representation (IDR) discriminative evaluation, qualitative and quantitative comparison of different methods on unseen degradations. 
Sec.\ref{visual} showcases the recovery effects of various methods across multiple degradation scenarios.
\vspace{-4pt}
\section{Additional Ablation Study}\label{add_ablation}
To comprehensively validate the effectiveness of the proposed method, we further analyze the effects of each component under \textit{Degradation Setting 2}.
We select nine anisotropic Gaussian blur kernels. For each kernel, we show the average PSNR on B100 benchmark \cite{b100_37} under degradation combinations with noise levels of 5 and 10.
\vspace{-2pt}
\subsection{Core Components for Training}
Similarly, we primarily analyze the effect of degradation reference prior (DRP), contrastive learning (CL), and the number of positive samples on SR performance.\par
\textbf{1) Effect of DRP and CL.} 
These two are used to enhance the discriminability of IDRs. The results are shown in Table \ref{table:x4_xr1}. 
We also build four models: 
T1 is the baseline model without DRP and CL. 
T2 incorporates DRP and achieves an average PSNR improvement over T1 of 0.20 dB across different degradation conditions. 
T3 introduces CL and achieves an average PSNR improvement over T1 of 0.17 dB across different blur kernels.
By combining DRP and CL, our method (T4) achieves 0.26 dB improvement in average PSNR on all blur kernels compared to T1.
The above results indicate that DRP and CL are crucial for training the IDR-estimator, and they are compatible.
\par
\begin{figure}[h] 
      {
      \centering
\centerline{\includegraphics[width=0.98\linewidth]{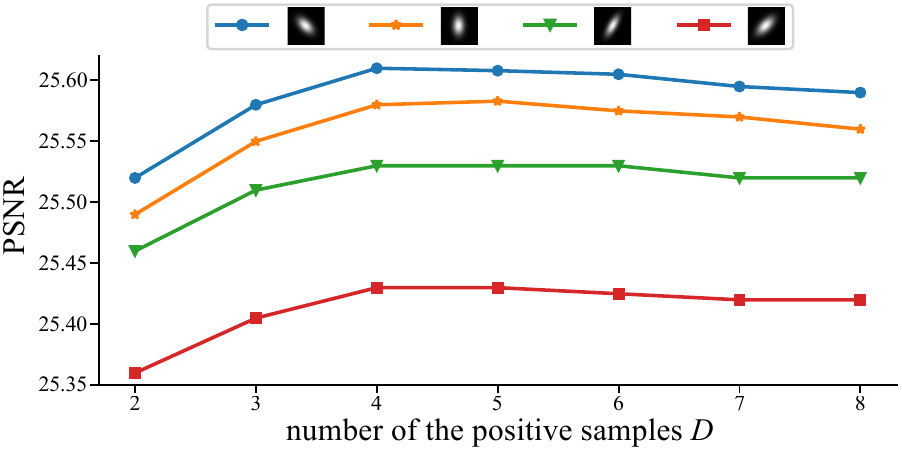}} 
      \vspace{-3pt}
      \caption{
      The effect of different numbers of positive samples.
      }
      \label{fig:postive}
      }
      \vspace{-15pt}
\end{figure}

\textbf{2) Impact of Positive Sample Number.} 
To further verify the impact of the number of positive samples $D$ on the SR performance, we conduct experimental analysis under four different anisotropic Gaussian blur kernels. The results are shown in Fig.\ref{fig:postive}. Consistent with the conclusion under the \textit{Degradation Setting 1}, our method achieves the best balance between performance and training computational cost when the number of positive samples $D$ is set to 4.
\vspace{-3pt}

\begin{table*}[h]
      \renewcommand{\baselinestretch}{1.1}
{\footnotesize\centerline{\tabcolsep=5.4pt\begin{tabular*}{\textwidth}{c|cc|ccccccccc}
        \bottomrule
         \multicolumn{1}{c|}{\multirow{1}[3]{*}{Teacher}} & \multicolumn{1}{c}{\multirow{1}[3]{*}{\shortstack{DRP}}} & \multicolumn{1}{c|}{\multirow{1}[3]{*}{\shortstack{CL}}} & \textbf{\begin{minipage}[b]{0.15\columnwidth}
      \centering
      \raisebox{-.55\height}{\includegraphics[width=0.40\linewidth]{./fig/1.png}} 
      \end{minipage}} & 
      \textbf{\begin{minipage}[b]{0.15\columnwidth}
      \centering
      \raisebox{-.55\height}{\includegraphics[width=0.40\linewidth]{./fig/2.png}} 
      \end{minipage}} & 
      \textbf{\begin{minipage}[b]{0.15\columnwidth}
            \centering
            \raisebox{-.55\height}{\includegraphics[width=0.40\linewidth]{./fig/3.png}} 
            \end{minipage}} & 
      \textbf{\begin{minipage}[b]{0.15\columnwidth}
      \centering
      \raisebox{-.55\height}{\includegraphics[width=0.40\linewidth]{./fig/4.png}} 
      \end{minipage}} & 
      \textbf{\begin{minipage}[b]{0.15\columnwidth}
            \centering
            \raisebox{-.55\height}{\includegraphics[width=0.40\linewidth]{./fig/5.png}} 
      \end{minipage}} & 
      \textbf{\begin{minipage}[b]{0.15\columnwidth}
            \centering
            \raisebox{-.55\height}{\includegraphics[width=0.40\linewidth]{./fig/6.png}} 
      \end{minipage}} & 
      \textbf{\begin{minipage}[b]{0.15\columnwidth}
            \centering
            \raisebox{-.55\height}{\includegraphics[width=0.40\linewidth]{./fig/7.png}} 
      \end{minipage}} & 
      \textbf{\begin{minipage}[b]{0.15\columnwidth}
            \centering
            \raisebox{-.55\height}{\includegraphics[width=0.40\linewidth]{./fig/8.png}} 
      \end{minipage}} & 
      \textbf{\begin{minipage}[b]{0.15\columnwidth}
            \centering
            \raisebox{-.55\height}{\includegraphics[width=0.40\linewidth]{./fig/9.png}} 
      \end{minipage}} \\ [0.28cm]
    \hline
    T1 & \multicolumn{1}{c}{} & \multicolumn{1}{c|}{} & 26.28 & 26.20 & 25.52 & 25.37 & 25.33 & 25.28 & 25.17 & 24.91 & 24.58 \\
    T2 & $\checkmark$     & \multicolumn{1}{c|}{} & 26.47 & 26.43 & 25.73 & 25.56 & 25.52 & 25.47 & 25.36 & 25.10 & 24.79 \\
    T3 & \multicolumn{1}{c}{} & $\checkmark$  & 26.49 & 26.40 & 25.74 & 25.51 & 25.50 & 25.44 & 25.31 & 24.99 & 24.77 \\
    T4(Ours) & $\checkmark$  & $\checkmark$   & \textbf{26.53} & \textbf{26.46} & \textbf{25.77} & \textbf{25.61} & \textbf{25.58} & \textbf{25.53} & \textbf{25.43} & \textbf{25.15} & \textbf{24.86} \\
      \toprule
\end{tabular*}}}
\vspace{-5pt}
\caption{The effect of different training framework designs on the average PSNR for noise levels of 5 and 10 under nine different anisotropic blur kernels in \textit{Degradation Setting 2}.}\label{table:x4_xr1}
\vspace{5pt}
      \renewcommand{\baselinestretch}{1.1}
{\footnotesize\centerline{\tabcolsep=3.8pt\begin{tabular*}{\textwidth}{cccc|ccccccccc}
        \bottomrule
        \multicolumn{1}{c|}{\multirow{1}[3]{*}{Method}} & \multicolumn{1}{c}{\multirow{1}[3]{*}{\shortstack{$\widehat{R}_{S}$}}} & \multicolumn{1}{c}{\multirow{1}[3]{*}{\shortstack{$\widehat{R}_{C}$}}} & \multicolumn{1}{c|}{\multirow{1}[3]{*}{IDR-CB}} & \textbf{\begin{minipage}[b]{0.15\columnwidth}
      \centering
      \raisebox{-.55\height}{\includegraphics[width=0.40\linewidth]{./fig/1.png}} 
      \end{minipage}} & 
      \textbf{\begin{minipage}[b]{0.15\columnwidth}
      \centering
      \raisebox{-.55\height}{\includegraphics[width=0.40\linewidth]{./fig/2.png}} 
      \end{minipage}} & 
      \textbf{\begin{minipage}[b]{0.15\columnwidth}
            \centering
            \raisebox{-.55\height}{\includegraphics[width=0.40\linewidth]{./fig/3.png}} 
            \end{minipage}} & 
      \textbf{\begin{minipage}[b]{0.15\columnwidth}
      \centering
      \raisebox{-.55\height}{\includegraphics[width=0.40\linewidth]{./fig/4.png}} 
      \end{minipage}} & 
      \textbf{\begin{minipage}[b]{0.15\columnwidth}
            \centering
            \raisebox{-.55\height}{\includegraphics[width=0.40\linewidth]{./fig/5.png}} 
      \end{minipage}} & 
      \textbf{\begin{minipage}[b]{0.15\columnwidth}
            \centering
            \raisebox{-.55\height}{\includegraphics[width=0.40\linewidth]{./fig/6.png}} 
      \end{minipage}} & 
      \textbf{\begin{minipage}[b]{0.15\columnwidth}
            \centering
            \raisebox{-.55\height}{\includegraphics[width=0.40\linewidth]{./fig/7.png}} 
      \end{minipage}} & 
      \textbf{\begin{minipage}[b]{0.15\columnwidth}
            \centering
            \raisebox{-.55\height}{\includegraphics[width=0.40\linewidth]{./fig/8.png}} 
      \end{minipage}} & 
      \textbf{\begin{minipage}[b]{0.15\columnwidth}
            \centering
            \raisebox{-.55\height}{\includegraphics[width=0.40\linewidth]{./fig/9.png}} 
      \end{minipage}} \\ [0.28cm]
      \hline
      \multicolumn{1}{c|}{M1} &       &       &       & 25.82 & 25.64 & 24.87 & 24.82 & 24.89 & 24.75 & 24.63 & 24.36 & 24.09 \\
    \multicolumn{1}{c|}{M2} & \multicolumn{1}{c}{$\checkmark$} &       & \multicolumn{1}{c|}{$\checkmark$} & 26.49 & 26.43 & 25.74 & 25.59 & 25.55 & 25.48 & 25.41 & 25.11 & 24.84 \\
    \multicolumn{1}{c|}{M3} &       & \multicolumn{1}{c}{$\checkmark$} & \multicolumn{1}{c|}{$\checkmark$} & 26.51 & 26.44 & 25.74 & 25.60 & 25.57 & 25.51 & 25.42 & 25.13 & 24.85 \\
    \multicolumn{1}{c|}{M4(Ours)} & \multicolumn{1}{c}{$\checkmark$} & \multicolumn{1}{c}{$\checkmark$} & \multicolumn{1}{c|}{$\checkmark$} & \textbf{26.53} & \textbf{26.46} & \textbf{25.77} & \textbf{25.61} & \textbf{25.58} & \textbf{25.53} & \textbf{25.43} & \textbf{25.15} & \textbf{24.86} \\
    \multicolumn{1}{c|}{M5} & \multicolumn{1}{c}{$\checkmark$} & \multicolumn{1}{c}{$\checkmark$} &       & 26.47 & 26.40 & 25.71 & 25.57 & 25.50 & 25.45 & 25.32 & 25.11 & 24.75 \\
    \hline
    \multicolumn{1}{c|}{M4-} & \multicolumn{3}{c|}{M4 with error IDRs} & 26.24 & 26.19 & 25.20 & 25.20 & 25.11 & 25.02 & 24.90 & 24.33 & 23.82 \\
     \multicolumn{1}{c|}{M5-} & \multicolumn{3}{c|}{M5 with error IDRs} & 26.20 & 26.18 & 25.18 & 25.02 & 24.98 & 24.89 & 24.69 & 24.23 & 23.72 \\
      \toprule
\end{tabular*}}}
\vspace{-4pt}
 \caption{The effect of different IDR-AM designs on the average PSNR for noise levels of 5 and 10 under nine different anisotropic blur kernels in \textit{Degradation Setting 2}.}\label{table:x4_xr2}
 \vspace{-5pt}
\end{table*}
\begin{table*}[t]
      \renewcommand{\baselinestretch}{1.0}
      {\footnotesize\centerline{\tabcolsep=5.9pt
      \begin{tabular*}{\textwidth}{ccccccccccccccc}
    \toprule
    \multirow{2}[2]{*}{Method} & \multirow{2}[2]{*}{Param.} & Dataset & \multicolumn{3}{c}{Set5} & \multicolumn{3}{c}{Set14} & \multicolumn{3}{c}{B100} & \multicolumn{3}{c}{Urban100} \\
\cmidrule(r){3-3} \cmidrule(r){4-6} \cmidrule(r){7-9} \cmidrule(r){10-12} \cmidrule(r){13-15}     &       & Kernel width & 4.8   & 5.0   & 5.2   & 4.8   & 5.0   & 5.2   & 4.8   & 5.0   & 5.2   & 4.8   & 5.0   & 5.2 \\
    \midrule
    \textbf{DAN}   & 4.3M  & CNN   & \textcolor[rgb]{ 1,  0,  0}{\textbf{25.95}} & \textcolor[rgb]{ 1,  0,  0}{\textbf{25.36}} & \textcolor[rgb]{ 1,  0,  0}{\textbf{24.93}} & 24.24  & 23.84  & 23.47  & \textcolor[rgb]{ 0,  .439,  .753}{\underline{24.52}} & \textcolor[rgb]{ 0,  .439,  .753}{\underline{24.23}} & \textcolor[rgb]{ 1,  0,  0}{\textbf{23.98}} & 21.73  & \textcolor[rgb]{ 0,  .439,  .753}{\underline{21.34}} & 21.04  \\
    \textbf{DASR}  & 5.8M  & CNN   & 25.77  & 25.23  & 24.83  & 24.24  & 23.85  & \textcolor[rgb]{ 1,  0,  0}{\textbf{23.51}}  & 24.45  & 24.18  & \textcolor[rgb]{ 0,  .439,  .753}{\underline{23.95}} & 21.66  & 21.32  & \textcolor[rgb]{ 0,  .439,  .753}{\underline{21.05}} \\
    \textbf{MRDA}  & 5.8M  & CNN   & \textcolor[rgb]{ 0,  .439,  .753}{\underline{25.90}} & 25.30  & 24.84 & 24.10  & 23.70  & 23.35  & 24.46  & 24.13  & 23.86  & 21.64  & 21.23  & 20.93  \\
    \textbf{KDSR}  & 5.8M  & CNN   & \textcolor[rgb]{ 0,  .439,  .753}{\underline{25.90}} & \textcolor[rgb]{ 0,  .439,  .753}{\underline{25.31}} & 24.84 & \textcolor[rgb]{ 1,  0,  0}{\textbf{24.38}} & \textcolor[rgb]{ 1,  0,  0}{\textbf{23.90}} & \textcolor[rgb]{ 0,  .439,  .753}{\underline{23.50}} & \textcolor[rgb]{ 0,  .439,  .753}{\underline{24.52}} & 24.20  & 23.92  & \textcolor[rgb]{ 0,  .439,  .753}{\underline{21.77}} & 21.33  & 21.00  \\
    \textbf{DSAT}  & 15.6M  & Transformer   & 25.54 & 25.07 & 24.71 & 24.08  & 23.71  & 23.40  & 24.51  & \textcolor[rgb]{ 1,  0,  0}{\textbf{24.24}}  & \textcolor[rgb]{ 1,  0,  0}{\textbf{23.98}}  & 21.63  & 21.29  & 21.00  \\
    \textbf{CDFormer\textsubscript{S}} & 11.9M & Transformer & 25.71  & 25.18  & 24.77  & 24.10  & 23.72  & 23.38  & 24.41  & 24.13  & 23.89  & 21.56  & 21.21  & 20.92  \\
    \textbf{LightBSR}  & \textcolor[rgb]{ 1,  0,  0}{\textbf{3.1M}}  & CNN   & \textcolor[rgb]{ 1,  0,  0}{\textbf{25.95}} & \textcolor[rgb]{ 1,  0,  0}{\textbf{25.36}} & \textcolor[rgb]{ 0,  .439,  .753}{\underline{24.92}} & \textcolor[rgb]{ 0,  .439,  .753}{\underline{24.30}} & \textcolor[rgb]{ 0,  .439,  .753}{\underline{23.88}} & \textcolor[rgb]{ 1,  0,  0}{\textbf{23.51}} & \textcolor[rgb]{ 1,  0,  0}{\textbf{24.54}} & \textcolor[rgb]{ 1,  0,  0}{\textbf{24.24}} & \textcolor[rgb]{ 1,  0,  0}{\textbf{23.98}} & \textcolor[rgb]{ 1,  0,  0}{\textbf{21.83}} & \textcolor[rgb]{ 1,  0,  0}{\textbf{21.40}} & \textcolor[rgb]{ 1,  0,  0}{\textbf{21.08}} \\
    \bottomrule
\end{tabular*}}}
\vspace{-4pt}
\caption{Quantitative comparison of PSNR metric for different methods under unseen degradations on $\times$4 SR. 
      The best and second-best results are highlighted in \textcolor[rgb]{ 1,  0,  0}{\textbf{red}} and \textcolor[rgb]{ 0,  .439,  .753}{\underline{blue}}, respectively.}
      \label{table:unseen}
      \vspace{-16pt}
\end{table*}

\subsection{Core Components of IDR-AM}
\vspace{-1pt}
For IDR-AM, we primarily validate the effectiveness of the IDR Adaptation Block (IDR-AB) and the IDR Correction Block (IDR-CB). The results are presented in Table \ref{table:x4_xr2}.
\par
\textbf{1) Effect of IDR-AB.} 
There are two degradation modulation branches: channel-wise ($\widehat{R}_{C}$) and spatial-wise ($\widehat{R}_{S}$). 
M1 represents the baseline model without any modulation branch. 
Compared to M1, 
by adding spatial-wise modulation (M2) or channel-wise modulation (M3), the average PSNR on all blur kernels are 0.75 dB and 0.77 dB higher than that of M1, respectively.
Furthermore, by applying both spatial- and channel-wise modulation, the PSNR results on all blur kernels can be further improved compared to M2 and M3.
The above results show the effectiveness and complementarity of the two modulation branches.
\par
\textbf{2) Effect of IDR-CB.}
Firstly, we obtain M5 by removing the IDR-CB from M4 (ours).
It can be observed that the average PSNR of M5 is 0.07 dB lower than M4 on all blur kernels, demonstrating the promoting effect of IDR-CB on SR performance.
Secondly, we add random values ranging from 0 to 1 to the output of the IDR-estimator for both the M4 and M5 to simulate incorrect IDR inputs, and define the results as M4- (with IDR-CB) and M5- (without IDR-CB), respectively.
We can see that M4- achieves higher PSNR values than M5- across all degradation scenarios, demonstrating that the IDR-CB can effectively mitigate the adverse effects of incorrect degradation estimation. 
\vspace{-4pt}

\section{Out-of-distribution (OOD) Evaluation}\label{unseen}
To validate the generalizability,
we evaluate the performance of all methods in degradation scenarios outside the training range, 
including three experimental perspectives: IDR distribution, quantitative and qualitative evaluation.
\par
\subsection{The OOD evaluations in Degradation Setting 1}
Under \textit{Degradation Setting 1}, we compare the SR performance of different methods from both quantitative and qualitative perspectives.
The results are as follows.
\par
\textbf{Quantitative evaluation.} 
We use three different isotropic blur kernel widths \{4.8, 5.0, 5.2\} to evaluate the SR performance of various methods across four classic benchmarks. The comparison results with CDFormer\textsubscript{S} \cite{CDFormer}, DSAT \cite{DSAT}, KDSR \cite{kdsr}, MRDA \cite{mrda_9}, DASR \cite{dasr_5}, and DAN \cite{dan_13} are presented in Table \ref{table:unseen}. 
It can be seen that our proposed LightBSR achieves the best or second-best results across all degradation scenarios while maintaining the minimum number of parameters among all competitors.
\par
\begin{figure*}[t] 
      {
      \centering
\centerline{\includegraphics[width=0.97\linewidth]{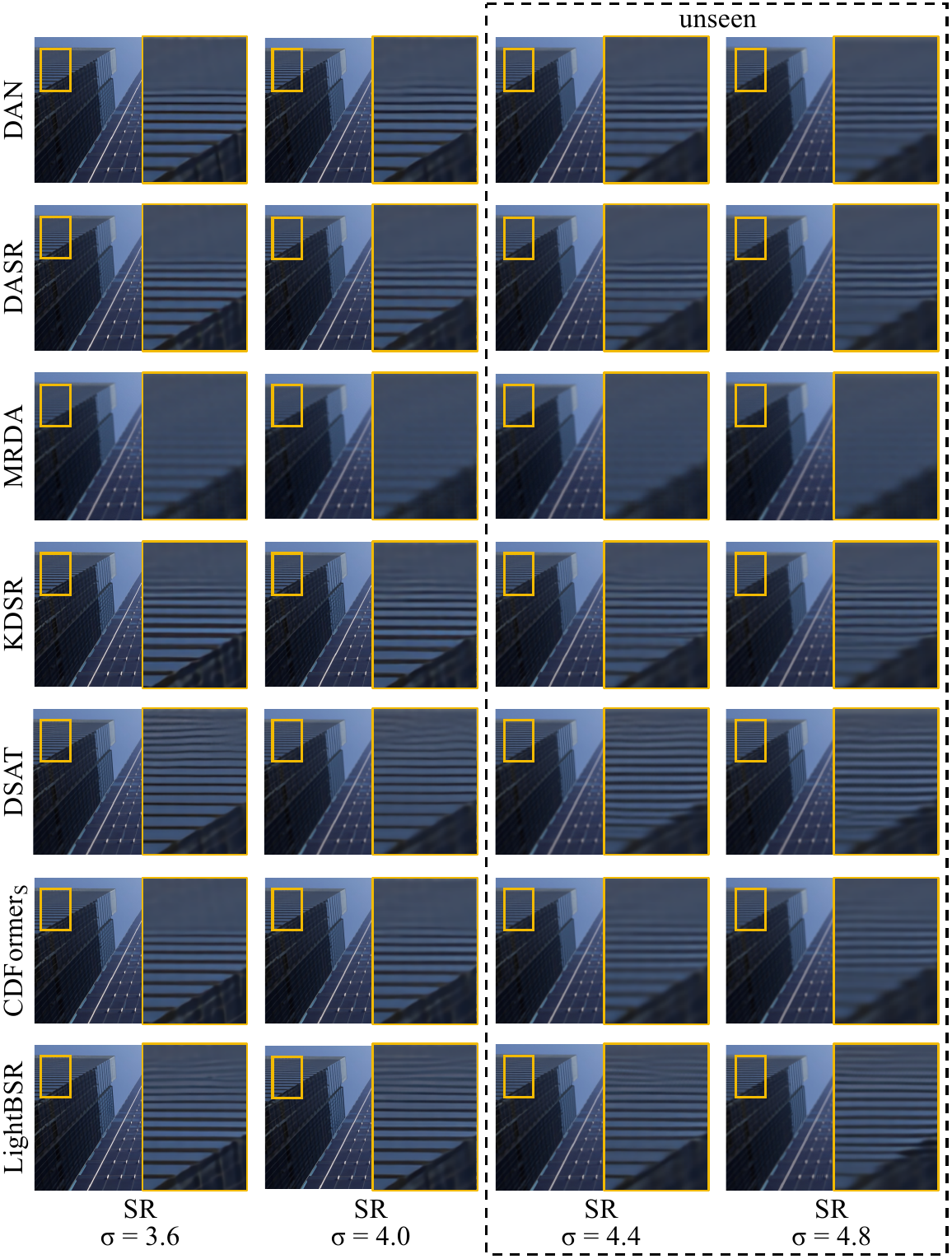}} 
      \caption{
      $\times$4 SR visual results of various methods on the ``img\_033'' of the Urban100 under unseen degradation settings. Zoom in to view additional details.
      }
      \label{fig:unseen}
      }
\end{figure*}

\textbf{Qualitative evaluation.} 
We also present the visual SR results of these methods on unseen degradations. The comparison results are shown in Fig.\ref{fig:unseen}. It can be observed that even when facing unseen degradation settings, LightBSR is capable of clearly and accurately restoring texture details, demonstrating the strong generalizability of our method.

\subsection{The OOD evaluations in Degradation Setting 2}
Under more complex \textit{Degradation Setting 2}, we focus on evaluating the discriminative ability of the IDR space learned by each method for OOD degradation patterns.
\par
\begin{figure}[t] 
      {
      \centering
\centerline{\includegraphics[width=0.99\linewidth]{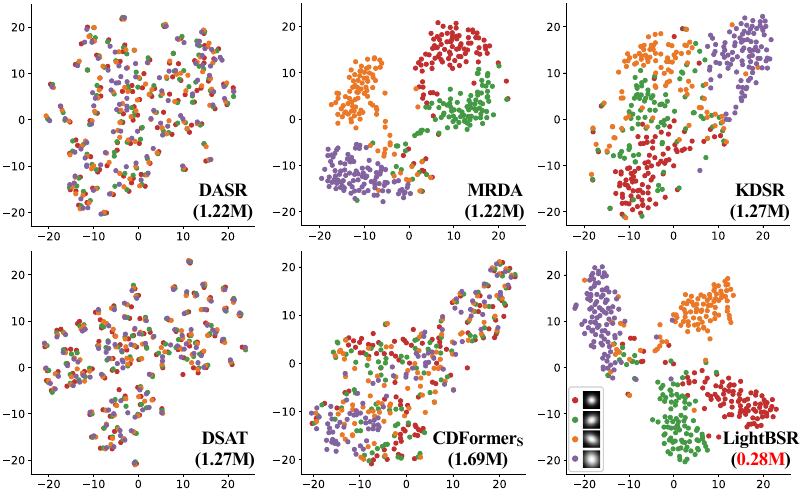}} 
      \caption{The t-SNE \cite{tsne_53} plots of IDR distributions on the B100 benchmark \cite{b100_37}. Four different unseen anisotropic blur kernels are chosen under \textit{Degradation Setting 2}, with the noise level set to 4.}
      \label{fig:s_tse}
      \vspace{-12pt}
      }
\end{figure}

\textbf{Quality of IDRs.} 
We apply four different unseen an-isotropic blur kernels on the B100 benchmark \cite{b100_37} to generate LR images, which are then input to DASR \cite{dasr_5}, MRDA \cite{mrda_9}, KDSR \cite{kdsr}, DSAT \cite{DSAT}, CDFormer\textsubscript{S} \cite{CDFormer} and LightBSR.
Comparing the IDR distributions of various methods in Fig.\ref{fig:s_tse}, 
it can be observed that:
1) CDFormer\textsubscript{S} struggles to differentiate different unseen degradations, demonstrating that diffusion-based estimator still needs optimization.
2) DSAT and DASR cannot deal with unknown degradation, indicating that only CL-based training framework cannot achieve good IDR generalization.
3) KDSR, MRDA and LightBSR all show the discriminability for various unseen degradations, with the latter two showing significant advantages.
These three methods all adopt the idea of KD and introduce degradation priors or multi-stage pipelines during training, showing that sophisticated IDR modeling is crucial for improving IDR generalization.
\par

\vspace{-4pt}
\section{Additional Visual Results}\label{visual}
\begin{figure*}[t] 
      {
      \centering
\centerline{\includegraphics[width=0.95\linewidth]{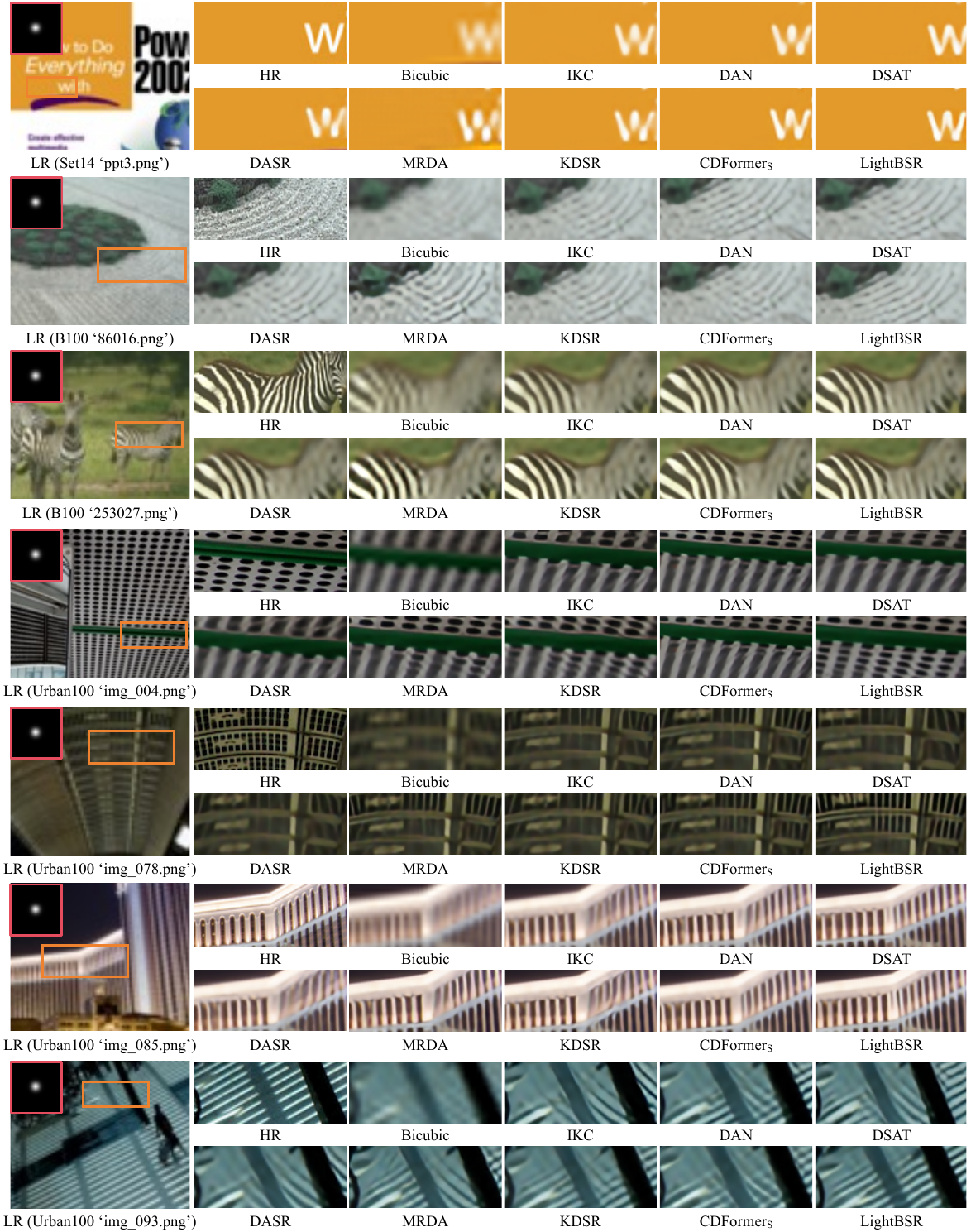}} 
      \caption{$\times$4 SR results for isotropic Gaussian blur kernel width 1.2 on Set14, B100 and Urban100 benchmarks.}
      \label{fig:x4_iso1.2}
      }
\end{figure*}

\begin{figure*}[t] 
      {
      \centering
\centerline{\includegraphics[width=0.95\linewidth]{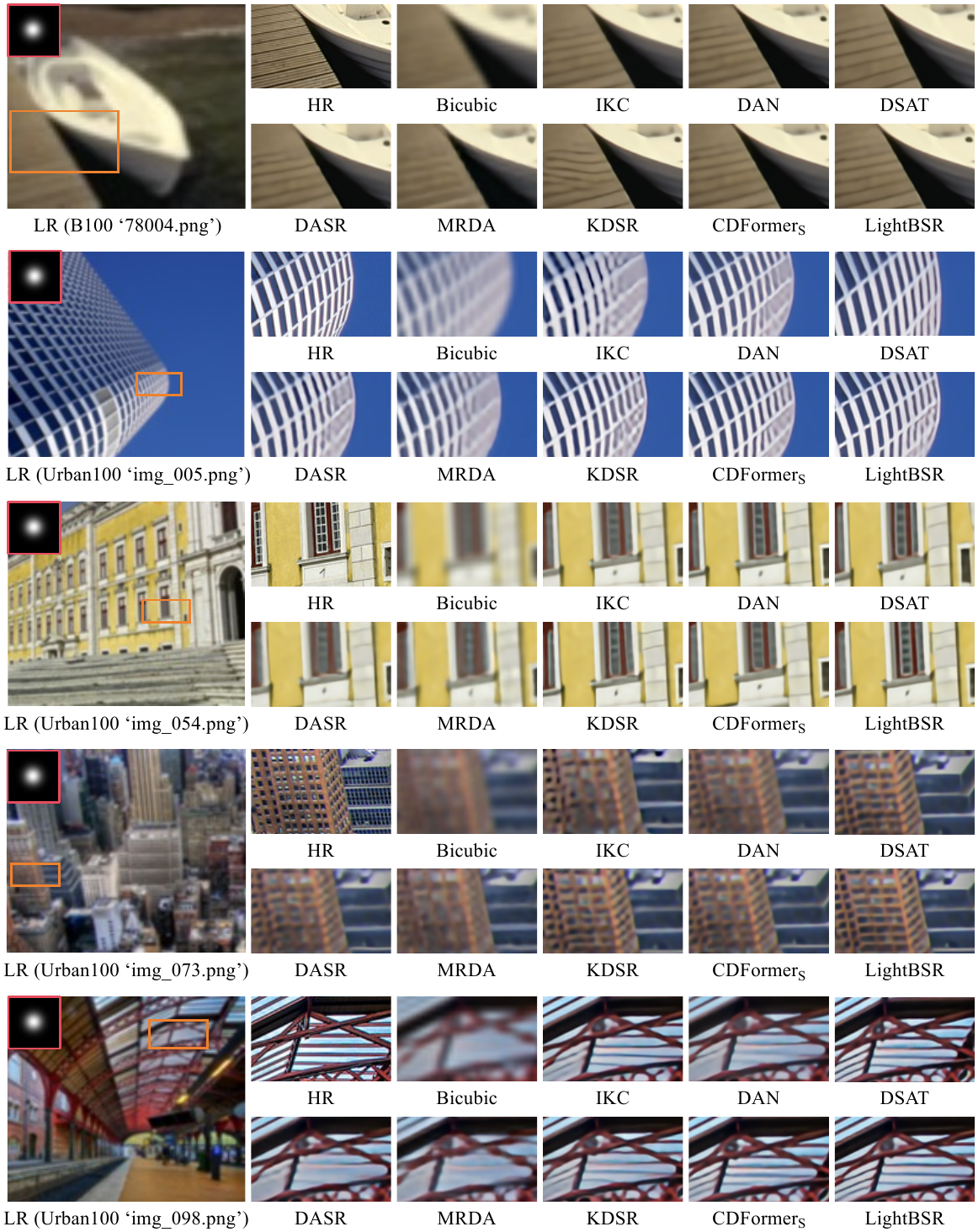}} 
      \caption{$\times$4 SR results for isotropic Gaussian blur kernel width 2.4 on B100 and Urban100 benchmarks.}
      \label{fig:x4_iso2.4}
      }
\end{figure*}

\begin{figure*}[t] 
      {
      \centering
\centerline{\includegraphics[width=0.95\linewidth]{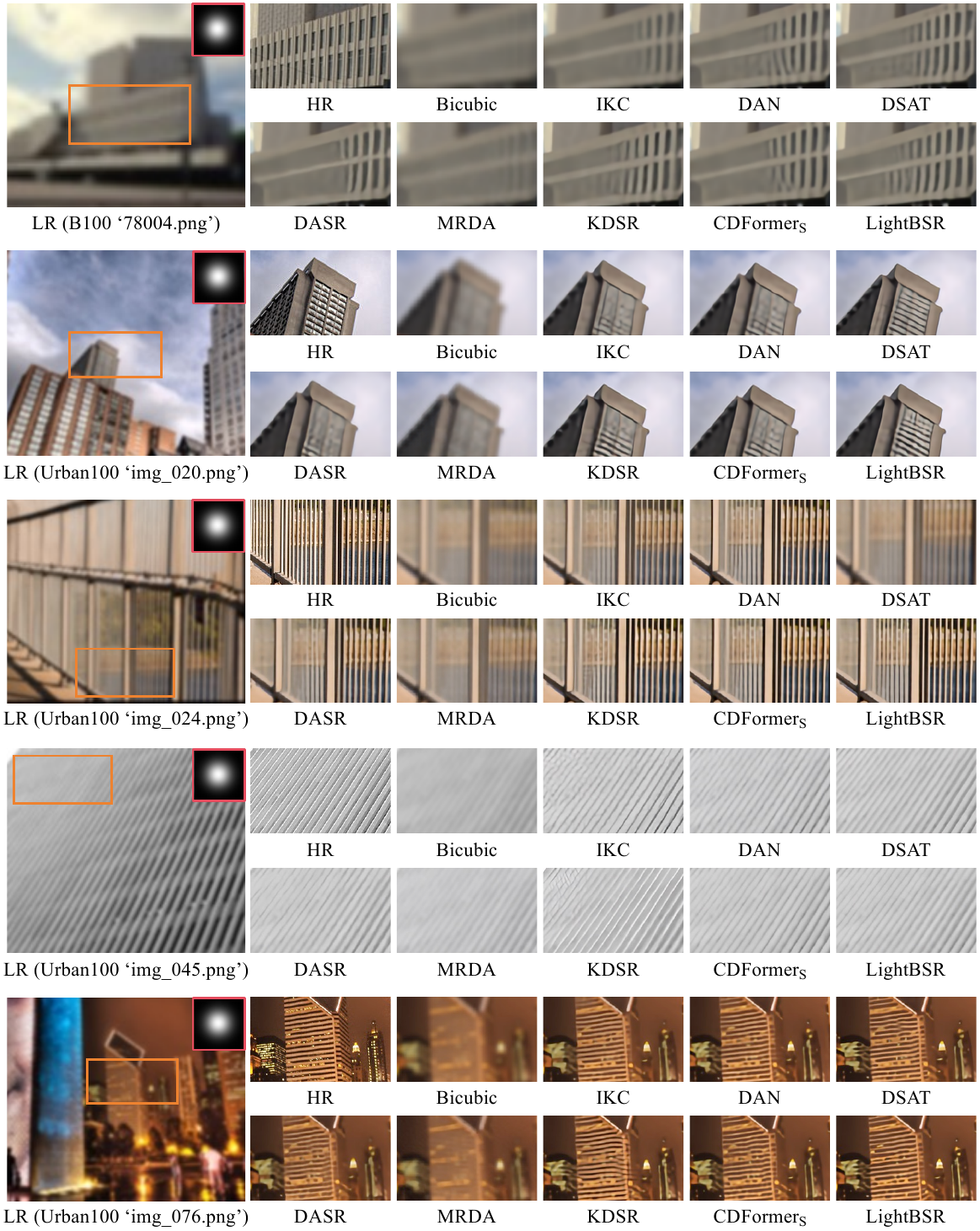}} 
      \caption{$\times$4 SR results for isotropic Gaussian blur kernel width 3.6 on B100 and Urban100 benchmarks.}
      \label{fig:x4_iso3.6}
      }
\end{figure*}

\vspace{-2pt}
\textbf{Visual Comparison in \textit{Degradation Setting 1}.} 
We present the $\times$4 SR results of different models under \textit{Degradation Setting 1} in Fig.\ref{fig:x4_iso1.2}, Fig.\ref{fig:x4_iso2.4}, and Fig.\ref{fig:x4_iso3.6}. 
The competitors include IDE-based methods CDFormer\textsubscript{S} \cite{CDFormer}, DSAT \cite{DSAT}, KDSR \cite{kdsr}, MRDA \cite{mrda_9} and DASR \cite{dasr_5}, and EDE-based methods IKC \cite{ikc_14} and DAN \cite{dan_13}. Compared to these methods, our method achieves superior visual restoration, producing clearer text, architectural details, and animal textures.
\par
\begin{figure*}[t] 
      {
      \centering
\centerline{\includegraphics[width=0.95\linewidth]{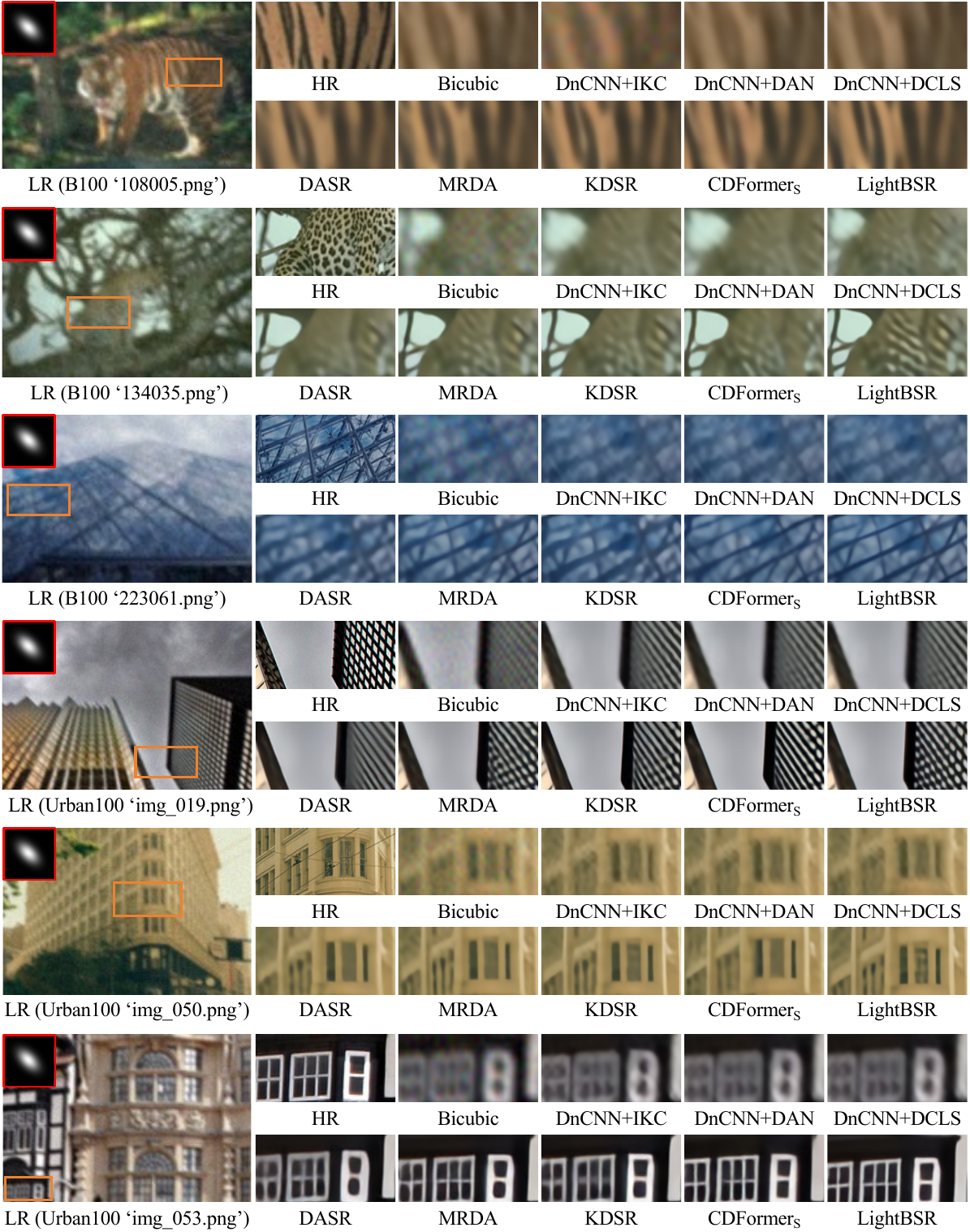}} 
      \caption{$\times$4 SR results on B100 and Urban100 benchmarks under an anisotropic Gaussian blur kernel with a noise level of 5.}
      \label{fig:x4_aniso_n5}
      }
\end{figure*}
\begin{figure*}[t] 
      {
      \centering
\centerline{\includegraphics[width=0.95\linewidth]{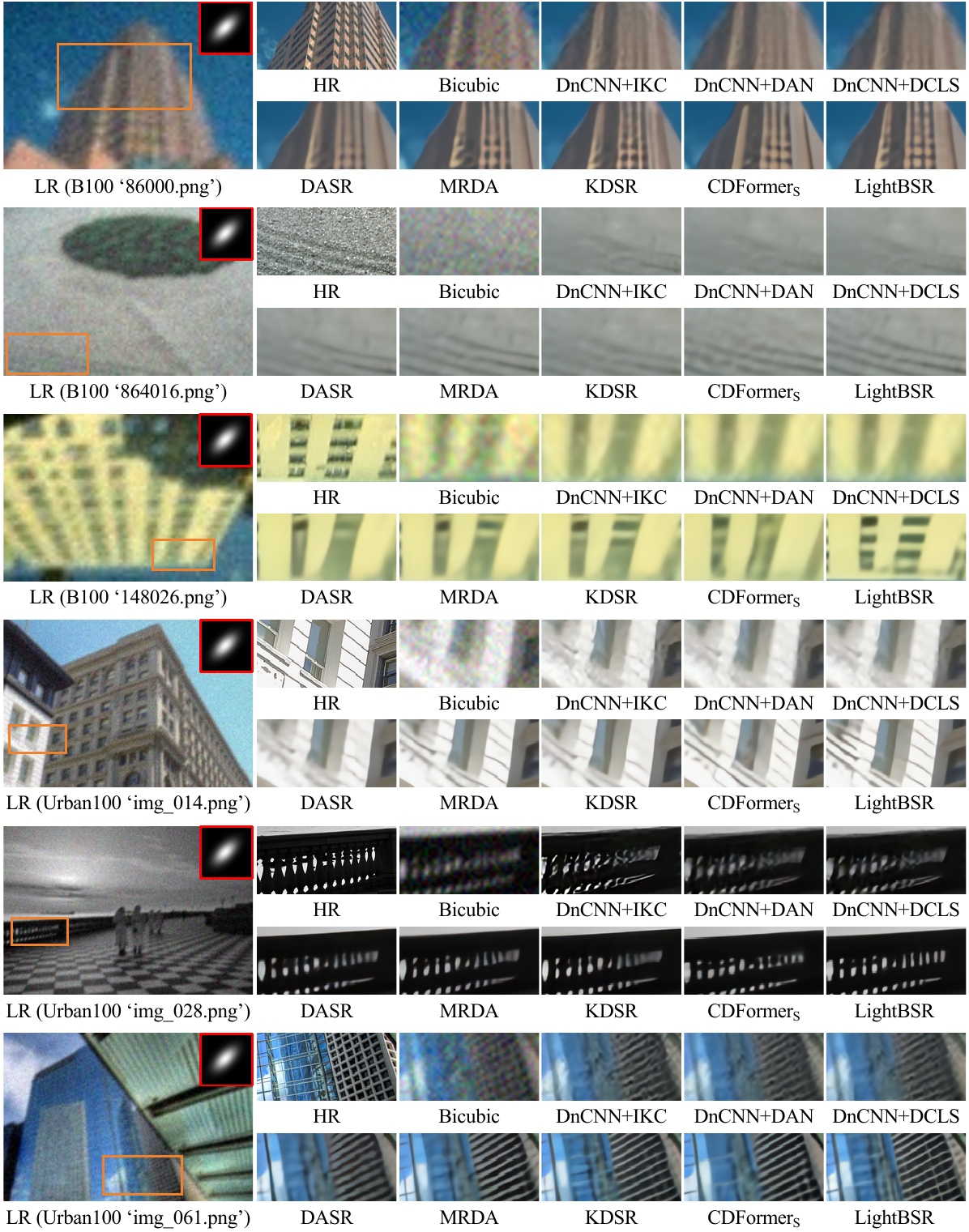}} 
      \caption{$\times$4 SR results on B100 and Urban100 benchmarks under an anisotropic Gaussian blur kernel with a noise level of 10.}
      \label{fig:x4_aniso_n10}
      }
\end{figure*}

\textbf{Visual Comparison in \textit{Degradation Setting 2}.} 
We provide the $\times$4 SR results under \textit{Degradation Setting 2} in Fig.\ref{fig:x4_aniso_n5} and Fig.\ref{fig:x4_aniso_n10}. 
Considering that IKC \cite{ikc_14}, DAN \cite{dan_13} and DCLS \cite{DCLS} are unable to denoise images, we first apply the DnCNN \cite{Dncnn} method as a denoising preprocessing step for these approaches. Compared to these methods, LightBSR produces clearer textures and more visually acceptable results under complex degradation scenarios.
\par
\begin{figure*}[t] 
      {
      \centering
\centerline{\includegraphics[width=0.95\linewidth]{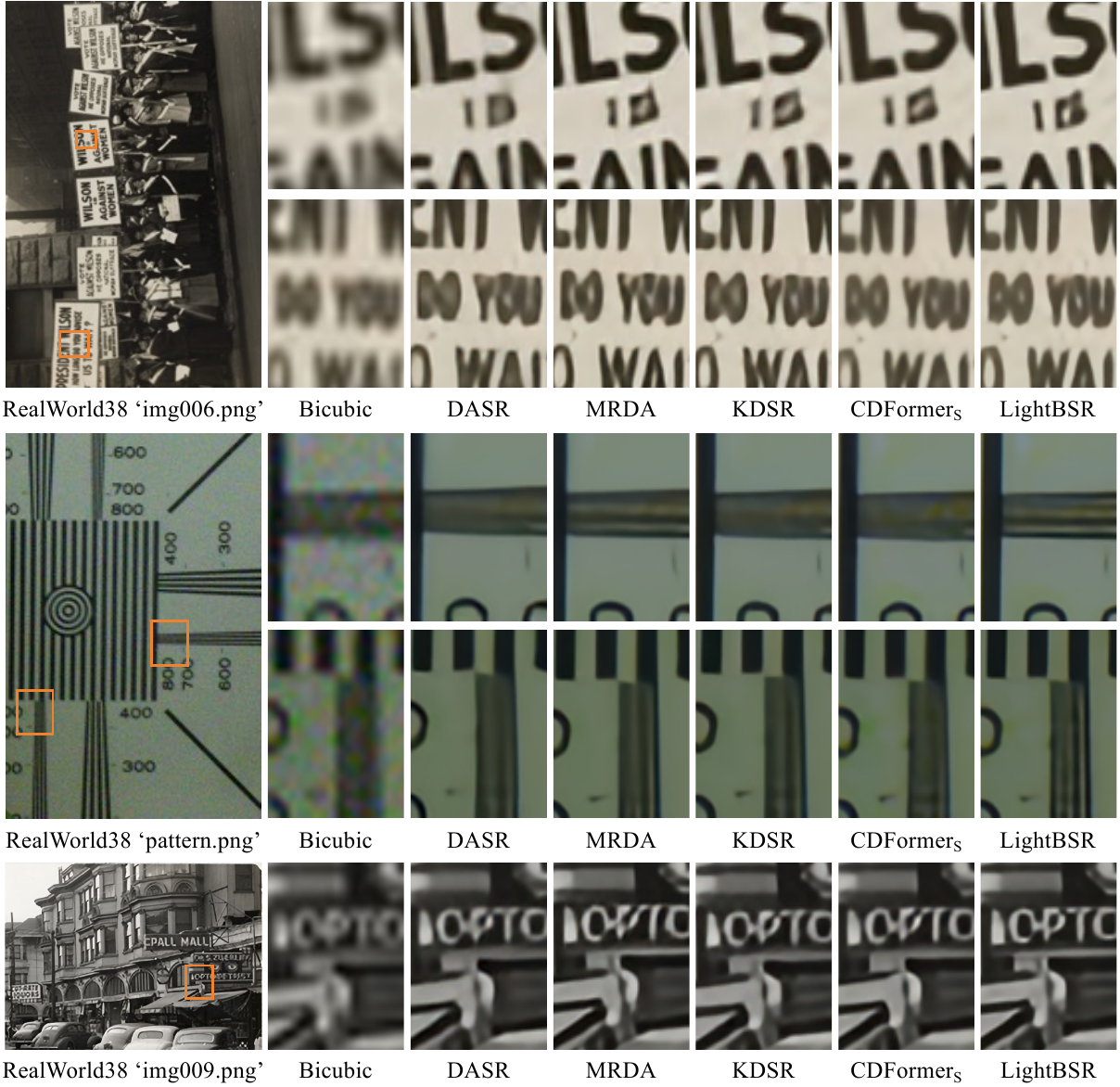}} 
      \caption{$\times$4 SR results on the RealWorld38 benchmark.}
      \label{fig:x4_realworld}
      }
\end{figure*}

\vspace{2pt}
\textbf{Visual Comparison in Real Degradation Scenario.} 
Finally, we also provide some $\times$4 SR results under real degradations, using the RealWorld38 \cite{dasrReal} dataset. The visual comparison of different models trained under \textit{Degradation Setting 2} is shown in Fig.\ref{fig:x4_realworld}. It can be observed that even for real degradations, our method also achieves satisfactory results in terms of detail and texture restoration.

\section{Related Work}\label{related work}
The key to BSR is to use a learnable estimator to extract degradation information from LR images instead of manually setting degradation parameters \cite{srmd_20,udvd}, to guide image reconstruction.
Early BSR methods explicitly predict degradation parameters, i.e. EDE-BSR task.
IKC \cite{ikc_14} iteratively refined the estimator by using generated SR images until the satisfactory SR result is achieved. 
DAN \cite{dan_13} introduced a two-branch network that predicts the blur kernel and SR image in parallel, alternately updating blur kernel estimation and SR reconstruction. 
DCLS \cite{DCLS} incorporated a deep constrained least square filtering module, adaptively producing deblurred features from the LR image.
Despite remarkable progress, this kind of method typically requires numerous iterations to compute degradation parameters, making it complex and time-consuming.
To solve this problem, some works focus on the implicit modeling of degradation, called IDE-BSR, with a research emphasis on learning a latent representation space for various degradations and integrating implicit degradation representation with LR features. 
DASR \cite{dasr_5} used contrastive learning \cite{mocov1} for the first time to model implicit representation of different degradations by learning the similarities and differences between samples.
IDMBSR \cite{idmbsr_7} incorporated the kernel width and noise level as weakly supervised signals to guide the implicit estimator training. 
MRDA \cite{mrda_9} used meta-learning \cite{meta-learning} and a multi-stage strategy to implicitly learn degradation representations.
KDSR \cite{kdsr} used the knowledge distillation (KD) \cite{kd,fitnets,ke2024rethinking} for the first time to learn the IDR estimator, where HR images are used to assist in teacher training, with the learned knowledge transferred to the student for degradation estimation during inference.
Recently, both DSAT \cite{DSAT} and CDFormer \cite{CDFormer} have adopted the design of building large SR networks by stacking Transformer blocks \cite{swinir_15,DAT} to achieve good effect, but over-complex models also limit their application.
Compared with latest methods that improve effect by expanding model parameters, our goal is to achieve a high-performance and low complexity BSR model by strengthening the discriminability of implicit degradation space.

\end{document}